\documentclass[10pt,a4paper]{article}

\usepackage{fullpage}
\usepackage{times}
\usepackage{epsfig}
\usepackage{graphicx}
\usepackage{amsmath}
\usepackage{amssymb}
\usepackage{xcolor}
\usepackage{url}
\usepackage[normalem]{ulem}
\usepackage{verbatim}
\usepackage{hyperref}

\usepackage{todonotes}

\usepackage{caption}

\newcommand{\be}{\begin{equation}}
\newcommand{\ee}{\end{equation}}

\newcommand{\bn}{\begin{eqnarray}}
\newcommand{\en}{\end{eqnarray}}

\newcommand{\ben}[1]{\begin{enumerate}#1\end{enumerate}}
\newcommand{\bi}[1]{\begin{itemize}#1\end{itemize}}

\newcommand{\bs}{\begin{split}}
\newcommand{\es}{\end{split}}

\newcommand{\RR}{\mathbb{R}}

\newcommand{\mc}{\mathcal}

\newcommand{\eqf}[1]{\begin{align}#1\end{align} }

\newcommand{\eq}[1]{\begin{align}#1\end{align} }
\newcommand{\eqn}[1]{\begin{align}#1\end{align} }

\begin{document}

%
%
%
%
%
%
%
%

\noindent\huge{\emph{Towards computational fluorescence microscopy: Machine learning-based integrated prediction of morphological and molecular tumor profiles}} \\[0cm]


{\centering{\large{Alexander Binder$^{1,6}$, Michael Bockmayr$^{2,10}$, Miriam H\"agele$^1$, Stephan Wienert$^2$, Daniel Heim$^2$, Katharina Hellweg$^3$, Albrecht Stenzinger$^4$, Laura Parlow$^2$, Jan Budczies$^2$, Benjamin Goeppert$^4$, Denise Treue$^2$, Manato Kotani$^5$, Masaru Ishii$^5$, Manfred Dietel$^2$, Andreas Hocke$^3$, Carsten Denkert$^{2,7}$, Klaus-Robert M{\"u}ller$^{1,8,9,*}$ and Frederick Klauschen$^{2,7,*}$\\[5mm]
}}}

\large{\noindent$^1$ Machine-Learning Group, Department of Software Engineering and Theoretical Computer Science, Technical University of Berlin, Marchstr. 23, 10587 Berlin, Germany\\
$^2$ Systems Pathology Lab, Institute of Pathology, Charit{\'e} University Hospital, Charit{\'e}platz 1, 10117 Berlin, Germany\\
$^3$ Department of Internal Medicine, Infectious Diseases and Pulmology\\ Charit{\'e}platz 1, 10117 Berlin, Germany\\
$^4$ Institute of Pathology, University of Heidelberg, Im Neuenheimer Feld 224, Heidelberg, Germany\\
$^5$ Laboratory of Cellular Dynamics, Immunology Frontier Research Center, Osaka University, 2-2 Yamada-oka, Suita, Osaka 565-0871, Japan\\
$^6$ ISTD Pillar, Singapore University of Technology and Design, 8 Somapah Road, 487372 Singapore\\
$^7$ German Cancer Consortium (DKTK), Charit{\'e}, Berlin Charit{\'e}platz 1, 10117 Berlin, Germany\\
$^8$ Department of Brain and Cognitive Engineering, Korea University, 145 Anam-ro, Seongbuk-gu, Seoul 136-713, South Korea\\
$^9$Max-Planck-Institute for Informatics, Saarbr\"ucken, Germany\\
$^{10}$ Department of Pediatric Hematology and Oncology University Medical Center Hamburg-Eppendorf, Hamburg, Germany\\
$^*$ Correspondence should be addressed to }

\thispagestyle{empty}
\newpage

\newpage
\tableofcontents

\begin{abstract}
Recent advances in cancer research largely rely on new developments in microscopic or molecular profiling techniques offering high level of detail with respect to either spatial or molecular features, but usually not both. Here, we present a novel machine learning-based computational approach that allows for the identification of morphological tissue features and the prediction of molecular properties from breast cancer imaging data. This integration of microanatomic information of tumors with complex molecular profiling data, including protein or gene expression, copy number variation, gene methylation and somatic mutations, provides a novel means to computationally score molecular markers with respect to their relevance to cancer and their spatial associations within the tumor microenvironment.
\end{abstract}

\section{Introduction}

While cancer research nowadays increasingly relies on molecular profiling based on "omics" technologies, it is becoming increasingly clear that certain aspects of tumor pathology need to be studied within tissue context\cite{beck_science}. This development is based on the insight that cancer progression is tightly linked to intercellular cross-talk and the interaction of the neoplastic cell with the surrounding microenvironment including the immune system\cite{pmid27481837}. While microscopic techniques allow to study biological processes with high spatial detail, relatively few molecular markers can be measured simultaneously microscopically despite recent advances\cite{Gerner2012,pmid24681723} in comparison to what is possible with “omics” analyses of bulk tissue. Moreover, the relationships between morphological tumor features and molecular profiling data with no or only very limited spatial resolution are largely unexplored. It is therefore desirable to find ways to bridge the gap between microscopic imaging approaches and high-dimensional "omics" technologies, thereby facilitating the discovery of localized molecular features that drive the spatially heterogeneous phenotype of a tumor.
	Here, we present a novel machine learning-based analysis approach that facilitates the integration of molecular and spatio-morphological information. More precisely, it allows for the spatial identification of morphological breast cancer tissue features and is at the same time capable of predicting molecular features including protein (PROT) and gene expression (RNASEQ), somatic mutations (SOM), copy number variations (CNV) and gene methylation (METH) and can also predict clinico-pathological parameters such as hormone receptor status, tumor grading and survival. The method is based on non-linear classifier decisions by layer-wise relevance propagation\cite{bach2015pixel}  capable of identifying with high spatial resolution cell and tissue properties, such as cancer cells or tumor infiltrating lymphocytes in data from different imaging modalities (brightfield, fluorescent/confocal). Moreover, training our method with image and molecular data from a novel large in-house database (B-CIB, Berlin Cancer Image Base) and The Cancer Genome Atlas (TCGA)\cite{TCGA2012}  facilitates the simultaneous prediction of multiple molecular features in breast cancer. From a set of over 60,000 tested molecular features (data on expression, methylation and copy number variations and mutations of ~20,000 genes as well as expression and phosphorylation of 190 proteins) we identify those molecular features that are predictable from morphological imaging data by our machine learning (ML) approach. In addition to predicting molecular features, this approach therefore allows to identify and rank molecular features with respect to their statistical association with tumor pathology. This information can be used to infer (and generate hypotheses on) their general relevance for (breast) cancer. 
	Moreover, similarly to how the approach not only classifies image content but derives the precise spatial locations of cells by assigning each image pixel a probability of belonging to a certain tissue/cell class (“cell type heat-mapping”), it can also be used to infer the spatial associations of morphological and molecular properties. This is possible even if trained with the typical non-spatially resolved ("bulk") molecular profiling data. Image regions down to individual pixels can then be assigned a score not only for a morphological property, but also for the respective predicted molecular feature. The resulting heatmap image is a computationally generated prediction of the spatial localization of the respective molecular properties (similar to a fluorescence microscopic image). Based on these data, correlations between molecular and morphological tissue properties can be estimated. Therefore, the computational method we propose contributes to integrating "omics" approaches with microscopic techniques by bringing closer together proteogenomic profiling of bulk tissue with (spatially-resolved) microscopic techniques. Because the approach is capable of estimating the contribution of certain spatial cell and tissue features to molecular predictions it may – similar to the above described prediction of the degree of the presence of a molecular feature – help infer or generate hypotheses on the relevance of different tissue components such as tumor-induced stroma or tumor infiltrating lymphocytes (TiLs) for certain molecular properties relevant to cancer.  
	Methodologically, the technique we present overcomes limitations of both classical segmentation-based image analysis techniques and previous machine learning algorithms\cite{DBLP:conf/miccai/AndreVBWA11,DBLP:journals/artmed/Cruz-RoaCG11,Yu2016}. While the former procedures often fail due to the high variance of structures to be segmented, the latter suffers from reduced resolution because an image region is classified as a whole without offering more precise spatially resolved information reflecting tumor heterogeneity. Our approach, which can be used with any cancer type, overcomes these limitations of both methods by back-tracing the contribution of the predicted properties to the precise spatial localization of morphological and molecular features ("computational fluorescence microscopy"). We test our novel concept by applying it to breast cancer imaging and molecular data to identify normal tissue, cancer cells, tumor infiltrating lymphocytes and tumor-induced stroma and to detect various molecular features using own data and those available at TCGA.

\section{Brief overview of methods}

\subsection{Machine learning-based prediction and spatial heatmapping}

Nonlinear machine learning methods are well established as predictors of medical and biological properties\cite{gurcanreview,schuffler_computational_2017,campanella_terabyte-scale_2018}. Until recently, only the prediction accuracy was of concern, whereas now also the nonlinear properties of the learning machine can be analyzed and made transparent\cite{bach2015pixel}. For a biological image analysis this corresponds to precisely mapping the results of the prediction for a novel, e.g. pathological slice onto a heatmap that reveals the morphological particularities of the respective pathological properties. The heatmapping algorithm consist of two parts. Firstly, forward prediction through the nonlinear learning machine where a classification between, say, cancer and no-cancer is made, rsp. a molecular property is predicted. Secondly, the prediction of the machine learning model is made transparent by backward propagation of the prediction score through the nonlinearities of the machine onto image features, namely, regions and pixels. Note that nonlinearity is typically required to obtain accurate predictions, e.g. the classifier is required to combine different pixels and or regions for example through kernel functions\cite{MueMikRaeTsuSch01}. 
The backward propagation (also denoted as layer-wise relevance propagation (LRP)) through the nonlinear predictor was done as described in Bach et al.5 for Bag-of-Words and other models and depends on the employed feature set in the forward prediction phase. We used three different feature sets for the three tasks of cancer patch classification, lymphocyte patch classification and whole slide protein/gene expression prediction (see supplement). 
Given the different application data (morphological, molecular and survival, see below), we employed Bag-of-Words features\cite{Csurka04} and kernel-based SVMs\cite{MueMikRaeTsuSch01,SonRaeHenWidBehZieBonBinGehFra10new} in order to avoid overfitting problems in such high dimensional-small sample problem regimes, that may challenge the usage of other techniques such as deep neural nets with interpretability extensions\cite{Lapuschkin_2016_CVPR}. 
	The Bag-of-Words feature of an image tile is computed in three steps. At first, a set of local features is computed over small patches within the tile. Then the local features are mapped onto visual words, which are points in the space of the local features. The visual words are obtained once during training by clustering local features from training images. Each mapping is a vector that has dimensionality equal to the number of visual words. Each dimension encodes a similarity to one visual word. 	Finally, the mappings of all local features are summed up and normalized to yield one Bag-of-Words feature vector for each tile. We used rank-weighted soft mapping\cite{DBLP:journals/cviu/BinderSMK13} in order to aggregate local features into the resulting Bag-of-Words representation. With respect to local features we used quantiles of norms of gradients and quantiles of image patch intensities in addition to the common color SIFT descriptors. This extension permits an improved discrimination between cancer and normal cells. See below and the supplement for more details on features.
	In a next step we compute different variants of SV-Kernels for the different application problems (morphological, molecular and survival) that subsequently enter the quadratic optimization and thus yield the final Support Vector Machine (SVM).  All used SV-Kernels are normalized to unit standard deviation in Hilbert space. Whenever a chi2-kernel was used, then the kernel width parameter was set as the mean of the chi2-distances between Bag-of-Words features extracted from a set of training images.
	As an overall note we would like to emphasize that all learning machines were trained according to best practice in machine learning, i.e. using cross-validation procedures for selecting a well generalizing model (e. g. \cite{Bishop:2006:PRM:1162264}). The features are extracted from the data using fixed parameter settings and are provided as inputs to the learning machine, i.e. no special adaptation or re-parametrization was performed.   

\subsection{Identification of tumor cells, lymphocytes and stroma}

From our image data base containing over 1,000 pathological images of tumor and normal tissue (at resolution 200x) and over 200,000 individually annotated cells, we used image patches of size 102x102 for training. The labels are binary for cancer and for lymphocytes, with positive label indicating presence of at least one cell of the respective type per patch. This so-called training data is then used for training of the learning machine (see above and supplement). 
	For predicting in out of sample data (or synonymously denoted as test data) whether cancer and lymphocyte is present, an image slide is subdivided into tiles of side length 102 with mutual overlaps between the tiles of 34 pixels. For each tile we computed one score (classification, regression) and one heatmap for cancer and lymphocyte detection using an SVM over a set of Bag-of-Words features using the respective SV-Kernels. A heatmap for the whole slide was obtained by appropriately averaging over all tile heatmaps, details can be found in the supplement. 
	For classification of cancer in histological slides we compute three Bag-of-Words features to reflect the a-priori knowledge on different morphological properties. The local Bag-of-Words feature are the concatenation of SIFT, 2x9 quantiles over norms of gradients and 2x9 quantiles over color intensities for one color channel. The local features are concatenated for the color channels red and blue amounting to (128+18+18) x (red, blue) = 328 dimensions. The local feature is extracted at 3 scales corresponding to a local feature radius of 9, 12 and 15, always with an overlap of 3 pixels. Each scale yields one bag of word feature with 510 dimensions resulting from 510 visual words. For cancer presence we use a sum of chi2-kernels over the 3 Bag-of-Words features. 
	For the quantiles over gradient norms used as local feature, we compute the set of all norms of gradients over one color channel for an image patch. The local feature patch is divided into two regions. In each region we compute 9 quantiles taken from 10\% to 90\%. This results in an 18-dimensional feature per color channel. For quantiles over intensities we employ analogously all pixel intensities for one image patch and color channel.
	For detection of lymphocytes we compute 6 Bag-of-Words features. These are defined by three types of local features combined with the two local feature radii of 9 and 12 pixels. The local features are SIFT (local feature dimension 128 per color channel, visual word size 510), quantiles over gradient norms as above (local feature dimension 18 per color channel, visual word size 384) and quantiles over image intensities (local feature dimension 18 per color channel, visual word size 384). Color channels are red and blue.
For lymphocytes we use a sum of histogram intersection kernels over the 6 Bag-of-Words features. For more details, also motivation the particular choices of features and SV-Kernels please see the supplement.

\subsection{Molecular property prediction}

Analogously to the morphological classification, we predict molecular properties by combining Bag-of-words features, SV-Kernels and an SVM. Given the complex molecular properties of biological systems and particularly cancer we transform the intrinsically more involved regression problem into a classification task (high versus low molecular property). This can alleviate the anticipated amount of measurement associated and biological noise as well as the small sample size in the prediction problem. The prediction was done for each molecular property separately.

\underline{Labels:}\\
The label for the protein expression prediction is binary, obtained by thresholding the protein expressions of the patient population for the selected protein. The threshold is chosen from the 9 equally-spaced quantiles between 10\% and 90\% of the values of the selected protein.  It accounts for the fact that for the majority of considered genes/proteins there is no known cutoff which  separates the patient population in groups with distinct biological behavior.

\underline{Features:}\\
The feature used for the prediction of protein expression states is computed as the appropriately averaged Bag-of-Words features over 102x102 patches spaced on a regular 34-stride grid over the patient image. In order to minimize the effect of stain variations, the Bag-of-Words feature is based on SIFT features only, computed over red and blue color channels, sampled on a grid of stride 3. The SIFT feature is computed on a square with a radius of 12 pixels. For the training data a visual vocabulary of 510 words is computed by k-means clustering, in line with standard practice\cite{DBLP:journals/cviu/BinderSMK13}. The final Bag-of-Words features are generated by mapping the SIFT features by rank-weighted soft mapping, summing the resulting mapping vectors and normalizing them to unit L1-norm, resulting in a 510 dimensional Bag-of-Words feature.

\underline{Learning algorithm:}\\
A support vector machine employed with a histogram intersection kernel is used to predict protein expression states from the averaged Bag-of-Words feature. The SV-kernel is normalized to have unit standard deviation in Hilbert space. This choice ensures that the optimal regularization constant is close to one in practice\cite{DBLP:conf/icml/ZienO07}. 

\underline{Experimental setup:}\\
Given the small sample size of less than 600, the experiment and the corresponding model selection procedure (where hyperparameters such as regularization are chosen) is run for each protein and quantile with 10-fold outer cross-validation and 9-fold inner cross-validation respectively. Thus, for each step of the outer cross validation, the best regularization constant is chosen by the error rate of the 9-fold inner cross-validation - which is performed on the training set of the same outer cross validation step. To ensure the validity of out-of-sample results, the test set of each step of the outer cross validation has never been touched to optimize parameters for the same step. 

The performance is reported on the test set of each outer-cross-validation step. The reported result is an average over the outer cross validation steps. Given the small sample size, choosing a single test set is likely to result in increased variance. Regularization constants for the cross-validation were taken between from 
	The performance reported is the average of true positive and true negative rate. In particular, the baseline under this measure obtained by a random or a constant prediction is always 50\%, even if we classify 10\% of the dataset against the remaining 90\%.
For each molecular target we report the performance measure as average of the test sets of the outer cross-validation split for the best performing quantile.The splits are randomized with the constraint, that for a k\% quantile as threshold, each split contains approximately k\% of the smaller expressions.

\underline{Statistical evaluation and multiple testing correction:}\\
To evaluate the statistical significance of our molecular predictions and correct for multiple testing / FDR-estimation we used an approach based on Hoeffdings's inequality\cite{hoeffdings} and Benjamini-Hochberg (BH)\cite{bhprocedure}. This approach tests the hypothesis that the computed balanced accuracy is significantly different from random prediction, corrects for multiple testing and identifies the molecular features that remain significant after FDR control. We perform an additional, complementary statistical analysis using Monte Carlo simulations to estimate the FDR by comparing true vs. random molecular label assignment (see supplement for a detailed description).

\underline{Molecular heatmapping:}\\
A heatmap for a molecular target level indicates localized evidence for high expression (ascending from green to yellow to red) versus evidence for low expression (ascending from green to blue). Neutral regions without any particular evidence are colored in green.

\subsection{Validation of molecular heatmapping}

We performed two different types of tests to validate the molecular heatmapping and show that the results are non-random and reflect actual biology. 
First, we performed immunohistochemical stainings of molecular tumor properties and compared them with the computational predictions for the same tumors. To this end, we predicted spatial protein patterns by computing molecular heatmaps for the DNA repair protein MSH2, the tumor suppressor p53 and the epithelial junction protein E-Cadherin based on Hematoxylin-Eosin (H\&E) stained images from in-house routine diagnostics in different breast cancers. Using consecutive slices of the same tumors we performed immunohistochemical stainings (IHC) for the respective proteins using standard routine laboratory procedures (see supplement for details) and merged the IHC results with image data of the Hematoxylin stained cell nuclei in which tumor cells were differentiated from lymphocytes and stromal fibroblasts and other normal cells by a pan-cytokeratin stain and controlled visually by a pathologist. The merged pseudo-color images (blue: IHC result, red: cancer cells, green: stromal cells and lymphocytes) were compared side by side with the original HE stained images and predicted molecular and morphological heatmaps. Computationally and experimentally generated images cannot be directly compared (pixel-wise or by overlaying the images) because the H\&E staining used for prediction and the immunohistochemical staining used for validation must be performed on separate tissue sections (see methods for detailed explanation). While subsequent tissue sections are similar, they often differ slightly due to morphological variations in 3D (i. e. along the axis perpendicular to the cutting plane) and minor technical distortions. And even slight deviations of only a few microns between computational and experimental images due to these effects would lead to "false-positive" differences. We therefore used the Quadrat test from geospatial statistics that computes and statistically evaluates the co-localization of different image properties in image regions\cite{gatrell1996}. This allows us to quantify and evaluate the significance of the spatial association of morphological and molecular properties in each image and to compare the ratios of image regions with and without co-localized morpho-molecular properties between computationally and experimentally generated images. With this approach we show, first, that the association of molecular and morphological properties is significant in both computational and experimental images and, second, that the patterns of the co-localization are similar in all pairs of computational and experimental image for each marker (see supplement for details).  
	As an additional test to demonstrate that the predicted spatial morpho-molecular associations are specific, we performed a control experiment based on synthetic data. To this end, we defined various geometric shapes representing different morphological cell types which were assigned synthetic molecular properties. Based on these definitions we generated synthetic histological images with similar distribution of property values as in the experimental data. Subsequently, we computed the molecular heatmaps in the same way as for the real histomorphological images and also evaluated their significance with the Quadrat test (see supplemental for details).

\subsection{Clinico-pathological data and survival analysis}
Clinical features hormone receptor status, tumor grade and overall survival were predicted in analogy to the molecular feature predictions. Differences in overall survival between different ML-predicted prognostic groups were computed using the log-rank test (see supplement).

\subsection{Imaging data and molecular tumor profiles}
Breast cancer samples for training and validation of morphological tissue properties and immunohistochemistry for the validation of the morpho-molecular heatmaps were obtained from the tissue archive of the Institute of Pathology, Charité University Hospital, from patients who had given their informed consent under the approval of the institutional review board.  
Additional imaging and molecular data were obtained from the publicly available TCGA data base through \url{https://tcga-data.nci.nih.gov}. CNV, RNASEQ, Protein and Methylation data were downloaded from GDAC. 
CNV: \url{http://gdac.broadinstitute.org/runs/analyses__2015_08_21/}  (TP.CopyNumber Gistic2 Level 4)
Protein data: \url{http://gdac.broadinstitute.org/runs/stddata__2015_11_01/data/} (MDAnderson level 3 protein normalization)
Methylation: \url{http://gdac.broadinstitute.org/runs/stddata__2015_11_01/} (Methylation preprocess level 3)

\subsection{Data availability: B-CIB image data base}

The training and test data for the learning machine are derived from this novel digital tissue data base B-CIB that comprises individually annotated cells of the following types: invasive-ductal breast cancer (invasive breast cancer of no special type according to latest WHO classification), invasive lobular breast cancer, normal breast gland epithelial cells, inflammatory cells (lymphocytes), stromal cells, endothelial cells. The data base is accessible at \url{http://www.b-cib.de} (website preview online, data will be made accessible upon publication).
The Berlin Cancer Image base contains over 200,000 pathologist-annotated cancer and normal cells including normal epithelial cells, tumor infiltrating lymphocytes, stromal and endothelial cells from over 1,000 breast cancers from routine diagnostics available on a tissue microarrays (TMA) and regular slides. The image data was digitized using virtual microscopy for brighfield and confocal scanning microscopy for fluorescence data, respectively. Based on the CognitionMaster framework\cite{Wienert2013CognitionMasterAO}, we developed a cell selection tool (the “Point-of-Interest (POI)-manager”), which facilitates a fast and precise manual labeling of cells in digitized histological images by trained experts. 

\subsection{GO enrichment and correlation analysis}
For the correlation analysis relating prediction scores and molecular signatures, the MSIGDB collection was used\cite{Subramanian15545}. The Gene Ontology (GO) collection C5 was replaced by up-to data downloaded from GO2MSIG\cite{Powell2014} using high quality annotations only (EXP,IDA,IEP,IGI,IMP,IPI,ISS,TAS) and gene sets containing 10 to 5000 genes. The Spearman correlation between the indicator functions of the 17264 categories and the prediction scores were computed. Cancer related signatures (33 for RNASEQ, 13 for CNV,15 for METH, 14 for SOM and 13 for PROT) were selected from the 500 terms with the strongest correlation for each data type separately. 
	To verify the statistical significance in the context of multiple testing, correlations between the selected categories and 1000 random permutations of the prediction scores were computed. Redundant categories were removed. Finally, 10 categories with correlation exceeding the 99 percentile (RNASEQ, CNV and METH) as well as 8 (PROT) and 5 (SOM) exceeding the 95 percentile in the permutation analysis were selected and represented in Fig.~\ref{fig:fig3mainms}.

For each selected signature, p-values were computed from the percentile rank of the correlation with the prediction score in the permutation analysis. To account for the different sizes of the signatures, binning of their indicator functions was performed (20 bins for RNASEQ, CNV and METH and 5 bins for PROT and SOM) prior to graphical representation. Similarly, the Spearman correlations after binning (r-bin = r) were computed between the binned indicator functions and the binned prediction scores to obtain more comparable and better interpretable correlation values. 
\section{Results}

\subsection{Identification of morphological cell and tissue features}

\begin{figure}[htp]
\includegraphics[width=\textwidth]{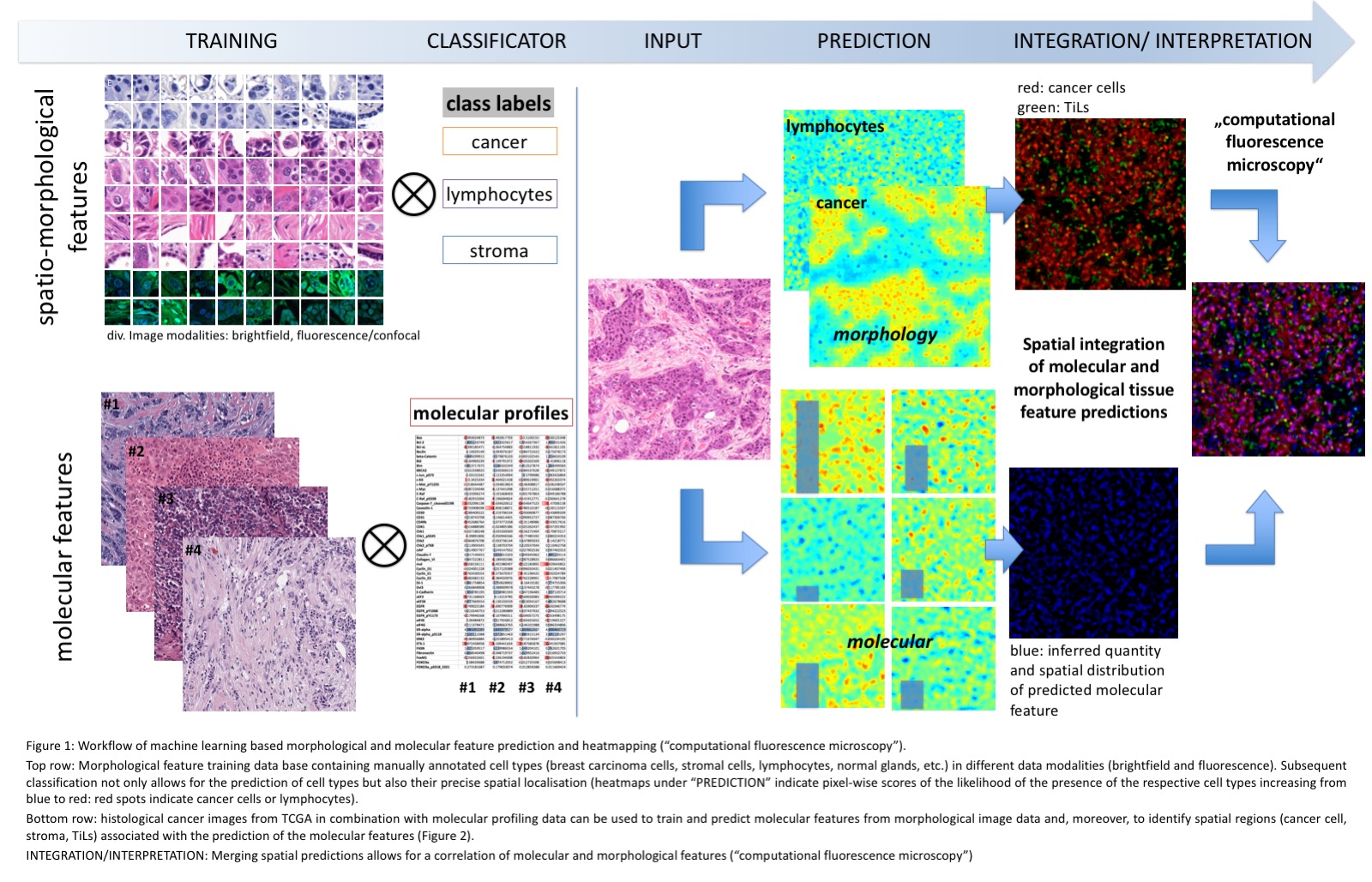}
\captionsetup{labelformat=empty}
\caption{\label{fig:fig1mainms} }
\end{figure}

\begin{figure}[htp]
\includegraphics[width=\textwidth]{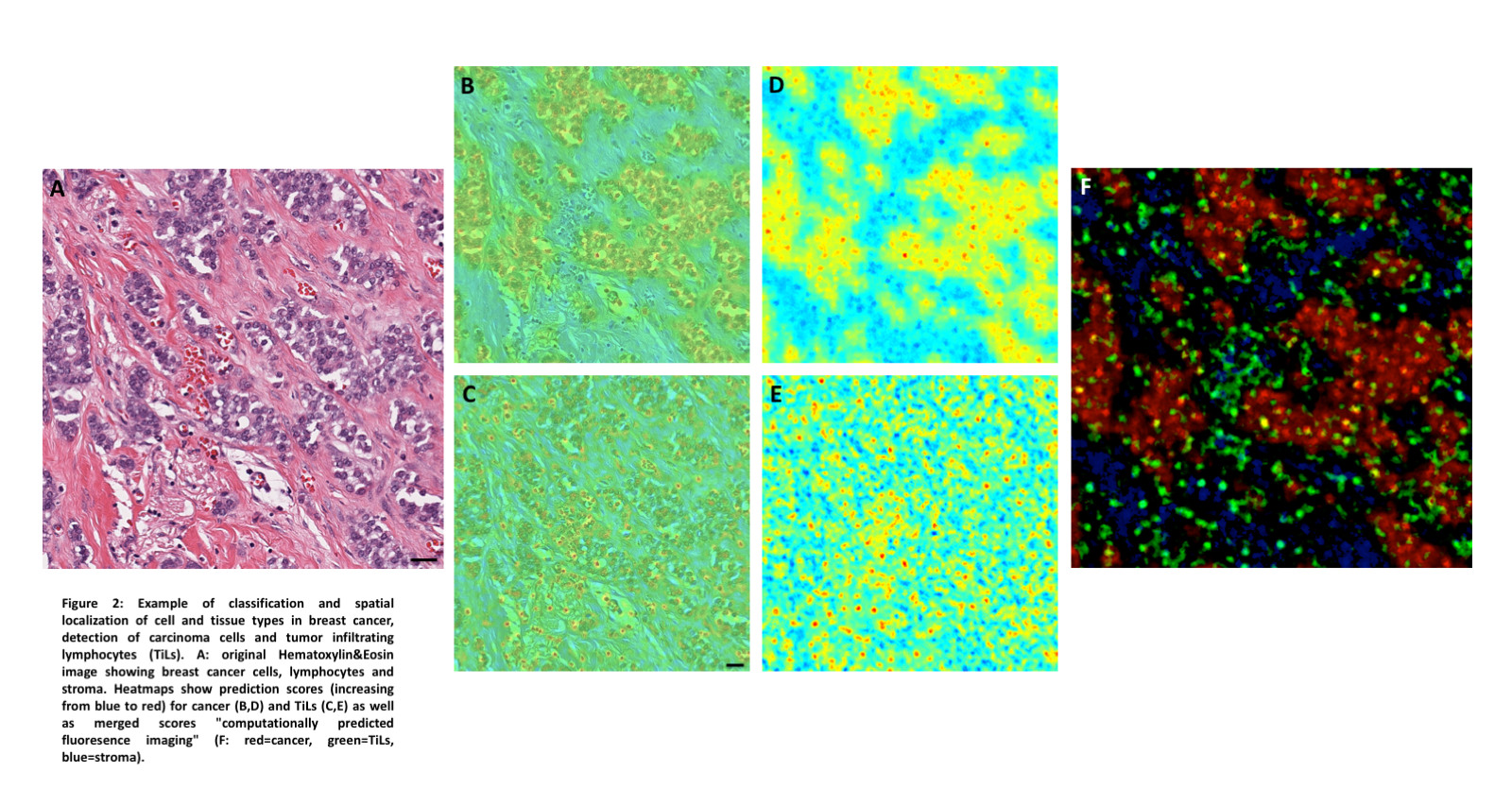}
\captionsetup{labelformat=empty}
\caption{\label{fig:fig2mainms} }
\end{figure}

Based on the different microscopic data sets we manually annotated different cell types in breast cancer tissue including a broad morphological spectrum of cancer cells as well as normal cells from stroma, vessels, glands and, in particular, tumor infiltrating lymphocytes. Overall, we created an image data base (Berlin Cancer Image Base, B-CIB) containing over 200,000 manually annotated cells from different imaging modalities comprising brightfield and fluorescence (confocal) microscopy data, which we use here to train our learning machine and which we make available to the community (see online methods for details). Morphological classification and localization performance shows best results for brightfield microscopy data with standard stains of over 95\% for specificity, sensitivity and average class wise accuracy, but only slightly lower values for confocal fluorescence image data (see supplement: Table \ref{tab:acc}). The better performance of conventional brightfield microscopy such as tissue stained with nuclear and cytoplasmic (hematoxylin \& eosin (H\&E)) and or nuclear (H) alone in comparison with confocal data stained for nuclei and cytoplasm is not surprising as the former dyes also stain the extracellular matrix and capture more morphological detail (see supplement: Fig. \ref{fig:suppl_spatfig} and \ref{fig:dist}). 
	Of note, for the present approach, training data for cell identification only needs to indicate the presence or absence of a particular cell type in the image patch facilitating a rapid manual annotation of training data. Moreover, this way of annotation is sufficiently flexible to not limit training to the cells (or the cell nuclei) themselves, but also to (implicitly) account for the fact that relevant information may also derive from extracellular context – even if no explicit manual annotation was performed for the microenvironment or molecular features5. Normal stromal or glandular tissue, cancer cells and lymphocytes can be robustly identified by this approach, which permits, for instance, the analysis of the spatial relationship between cancer cells and lymphocytes and could therefore be utilized for detection of tumor infiltrating lymphocytes (TiLs).
	For immediate visual validation and downstream analysis classification results are reconstructed as heatmaps showing per-pixel scores that reflect the likelihoods for the respective classification targets (Fig. \ref{fig:fig1mainms},\ref{fig:fig2mainms}). When overlaid with the histological image information these heatmaps offer a unique representation of the structural as well as quantitative features and help identify complex patterns of morphological tissue features complemented by the color shades indicating the likelihood/uncertainty of the classification results. We applied our approach to different stainings/image modalities (see supplement: Fig. \ref{fig:suppl_spatfig}) as well as immunohistological microscopy data showing the capability to combine detection of cancer cells in a complex background of stroma and normal epithelial components and quantification of nuclear immunohistochemical markers (such as the proliferation marker Ki67, supplement: Fig. \ref{fig:suppl_spatfig}: D,I,N).

\begin{figure}[htp]
\includegraphics[width=\textwidth]{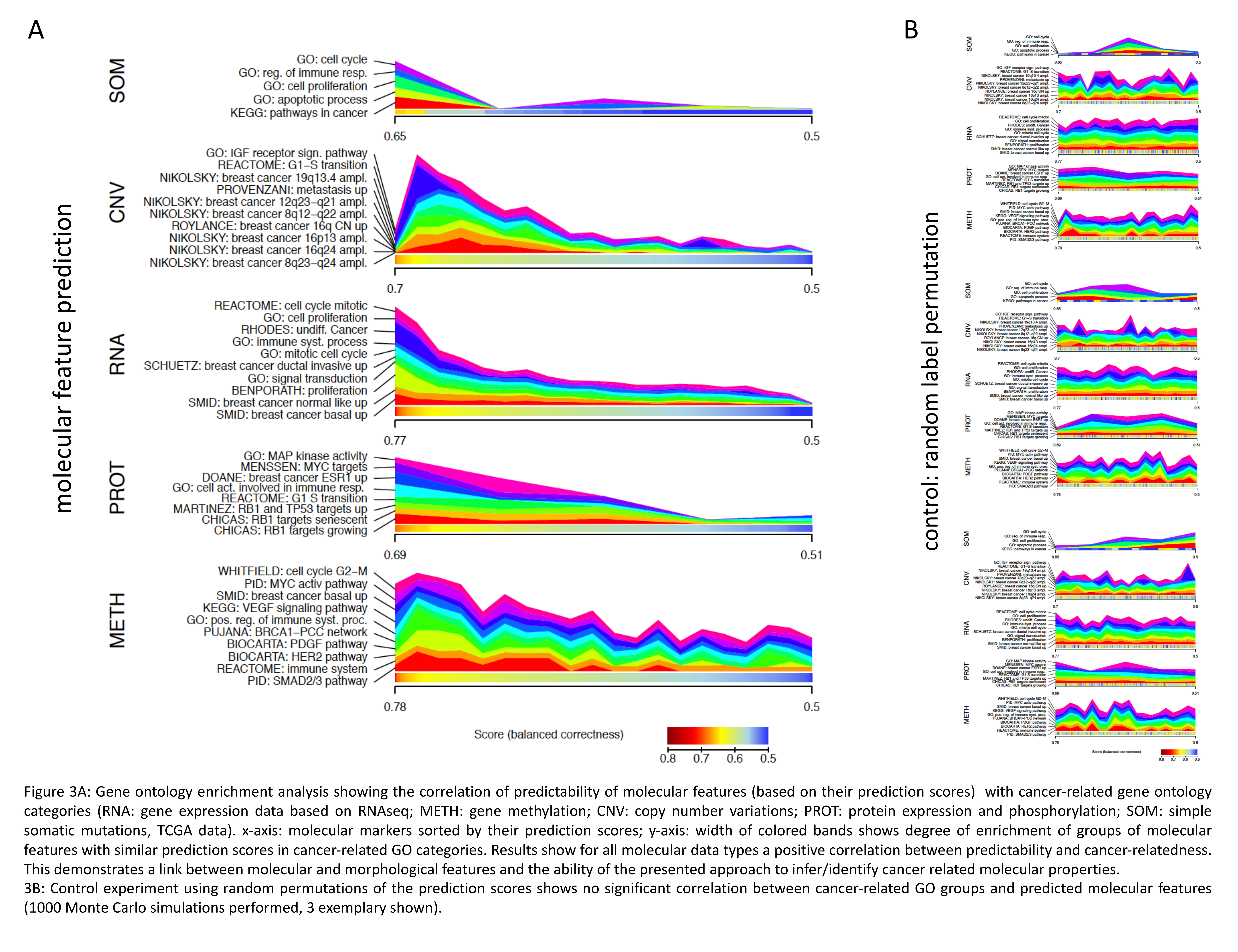}
\captionsetup{labelformat=empty}
\caption{\label{fig:fig3mainms} }
\end{figure}

\subsection{Prediction of molecular features from microanatomic image data}

While the classification and spatial localization of morphological cell and tissue features as described above is the basis of downstream analyses that can be used to characterize the different cellular components of the tumor microenvironment, we further extend our approach to also predict molecular features based on morphological image information. To this end, we combine our in-house data with histological and molecular profiling results of The Cancer Genome Atlas breast cancer data set, including data on all available simple somatic mutations, gene copy number variations, gene expression and methylation as well as protein expression and phosphorylation. Similar to the classification into different cell types, where training does not require precise spatial object annotation, locational information can be computationally deduced from the classification result; this is visualized as heatmaps through the layer-wise relevance propagation (LRP) procedure (see supplement and online methods). Therefore, training for molecular feature classification relies only on image data of representative tumor regions and "bulk" molecular profiles without requiring spatial information. 
	With the overall 565 cases, for which both image and molecular profile data were available, our approach is capable of predicting molecular properties with statistical significance corrected for multiple testing as follows: 
	 For simple somatic mutations, it was not feasible to analyze all genes, because only 23 genes were mutated in more than 10 cases in the available breast cancer cohort, out of which CDH1/E-Cadherin (1st) and TP53 (2nd) had the highest prediction scores (CDH1 $c=0.664$, $p=3.5e-04$; TP53 $c=0.62$, $p=9.1e-07$, significance analysis and multiple testing correction performed according to Hoeffding and Benjamini-Hochberg (BH), FDR-level $\alpha=0.05$) followed by GATA3 ($c=0.6$, $p=0.013$, $\alpha=0.05$) and PIK3CA ($c=0.59$, $p=2.9e-04$, $\alpha=0.05$). E-Cadherin mutations are known to be enriched in lobular breast cancer and its high predictive scores can therefore be seen as a validation of the method, because lobular carcinomas normally have a distinct morphology. 
	Out of the 190 proteins for which expression and phosphorylation was available ($n=565$ patients), our approach is capable of statistically significantly predicting 19 proteins with balanced accuracies up to $0.65$ (all $p<0.005$, $\alpha=0.05$), among which PR and HER2 are prominent markers used to identify clinically relevant subsets of breast cancer and E-Cadherin to differentiate between lobular and invasive ductal (no-special type) breast cancer. Other cell cycle or growth factor signaling-associated proteins predictable from morphology are known to be prognostic and predictive of therapy response in breast cancer (cyclin-B1 as a pharmacodynamic predictor\cite{Sabbaghi7006}, c-kit as a potential therapeutic target\cite{pmid29176653,pmid22711708}, FOXO3a is prognostic\cite{pmid23967095}).  
	Out of the 24,775 genes with available copy number variation data 5,274 genes (555 patients) were predictable with statistical significance by our approach with balanced accuracies between $c=0.6$ ($p<0.019$) and $c=0.7$ ($p=3.1e-17$; corrected for multiple testing with FDR-level  $\alpha=0.05$). For gene expression, we tested 20,530 genes and found 7,076 to be predictable (563 patients) with balanced accuracies between $c=0.6$ ($p<0.018$) and $c=0.76$ ($p=3.9e-36$; FDR-level  $\alpha=0.05$). DNA methylation was predictable with statistical significance for 5946 out of 19,953 genes with balanced accuracies between $c=0.6$ ($p<0.022$) and $c=0.78$ ($p=9.1e-11$; FDR-level  $\alpha=0.05$; $n=400$ patients). Additionally performed Monte-Carlo-based FDR estimation yielded equivalent results (see supplement). Molecular features with higher prediction scores were enriched for cancer-related processes (see below). For an in-depth description of the applied statistical tests see supplemental text. A detailed list of results over all molecular features is available from the authors.
	The observed maximum prediction scores increasing from $c=0.65$ for protein data, $c=0.66$ for somatic mutations, $c=0.70$ for CNV to $c=0.77$ for gene expression data and $c=0.78$ for gene methylation indicate a higher association of gene expression and methylation with morphological tissue properties than for the other molecular features. Ranking all $\sim 60,000$ molecular features based on their prediction scores, we used gene ontology (GO) information to test whether those markers showing better predictability are known to be associated with cancer (compared to those features that could not be predicted by our approach). Using different sets of gene ontology data that are related to cancer (growth processes, signal transduction, immune system and DNA repair), the results show for gene expression, DNA methylation, copy number variations as well as for protein expression and phosphorylation and somatic mutations that molecular features known to be relevant for cancer and associated immunological processes can be predicted with higher balanced accuracy compared to less relevant features. The correlations between cancer-related signatures and predictability/association with morphology for over 60,000 tested molecular features are strongest for gene expression data (all $r_{RNASEQ}>0.79$, all $p<0.001$) and slightly lower, but still significant for CNV (all $r_{CNV}>0.66$, all $p<0.003$), DNA methylation and somatic mutations (SOM) (all $r_{METH}>0.45$, $p<0.008$; $r_{SOM}>0.45$, $p<0.036$) and lowest for protein data (all $r_{PROT}>0.31$, $p<0.033$; p-values obtained from random subsampling, for more details on method see supplement.). While the protein data are biased due to the selection of only about 190 proteins known to be relevant for cancer, and the somatic mutation data have a limited base because only 23 genes are mutated in more than 10 cases in the cohort, they match the general observation made for the genome-scale data that molecular features relevant for cancer have higher prediction scores (Fig. \ref{fig:fig3mainms}). A detailed list of results over all molecular features is available from the authors. 

\subsection{Prediction of clinico-pathological tumor properties}

To evaluate to what extent clinically relevant tumor properties can be predicted from breast cancer morphology, we trained and predicted hormone receptor status, tumor grading and overall survival. Hormone receptor (HR) status is predictable with a balanced accuracy of $c=0.64$ ($p=4e-05$). While this result is highly significant, the overall average accuracy would be insufficient for clinical routine application. However, support vector machine (SVM) scores can be used to identify subsets of patients with substantially higher accuracies: SVM scores of $s \ge 0.78$ allow to identify as HR positive 37\% of the patients with an accuracy of 90\% or for $s\ge 1.3$ 13\% of the patients with an accuracy of 98\%. Similar, for tumor grade, differentiating between G1/2 and G3 balanced accuracy was $c=0.64$ ($p<3.7e-07$), but accuracies of over 93\% for correct G1/2 vs. G3 classification could be achieved with $s\le -1.2$ or $s\ge 1.04$ for 13\% of the patients. Feature prediction for survival analysis allowed us to identify two prognostic groups with mean overall survival $=$ 41.4 months vs. 47.2 months, $p=6e-10$; see supplement chapter \ref{sec:survtimeanalysis_new2018} for details on Kaplan-Meier statistics). We also attempted to predict clinical HER2-status, which was, however, not significant ($p=0.46$), demonstrating the differential impact different molecular properties have on morphology.

\subsection{Prediction of spatial localization of molecular features}

\begin{figure}[htp]
\includegraphics[width=\textwidth]{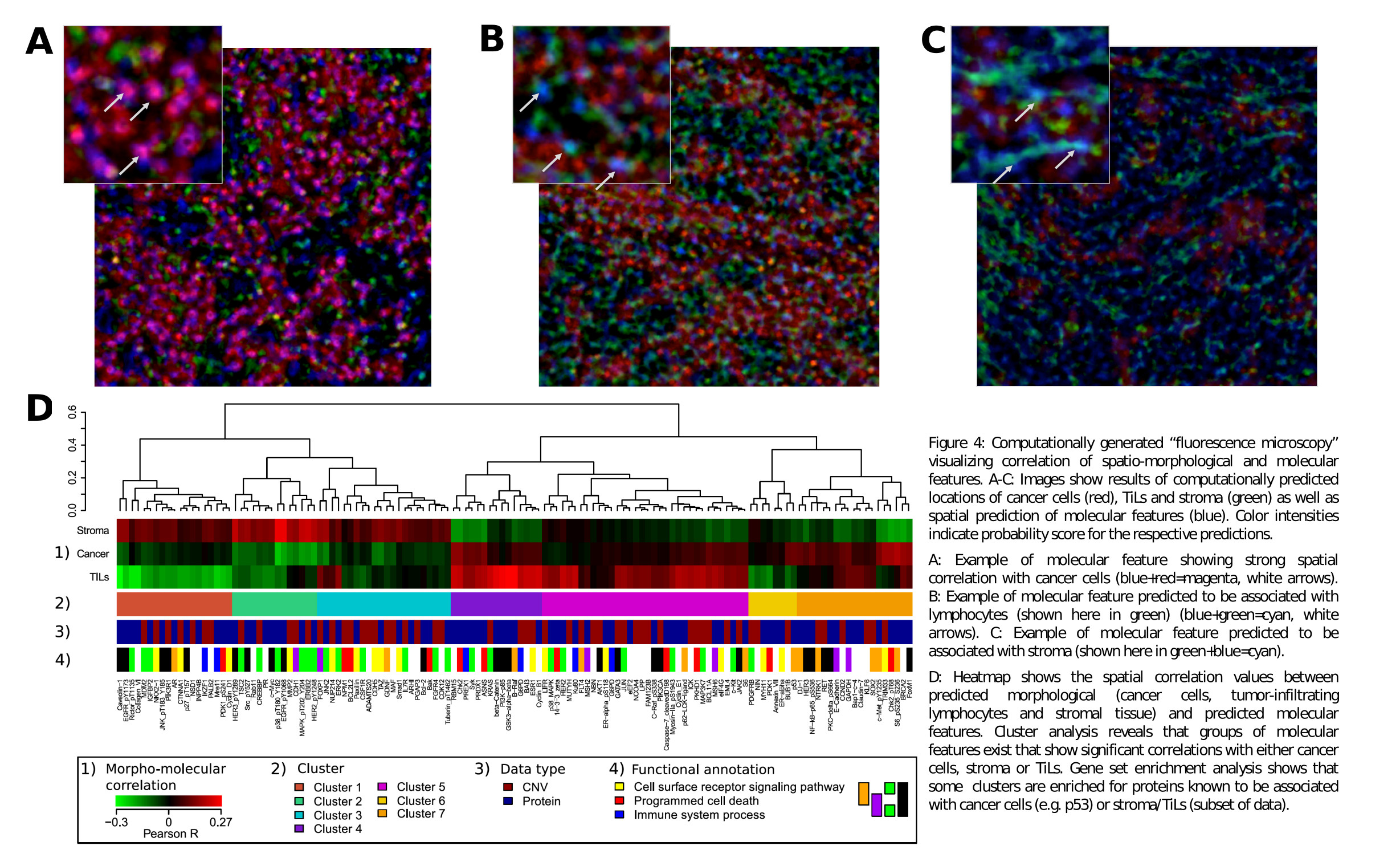}
\captionsetup{labelformat=empty}
\caption{\label{fig:fig4mainms} }
\end{figure}

\begin{figure}[htp]
\begin{center}
\includegraphics[height=\textheight]{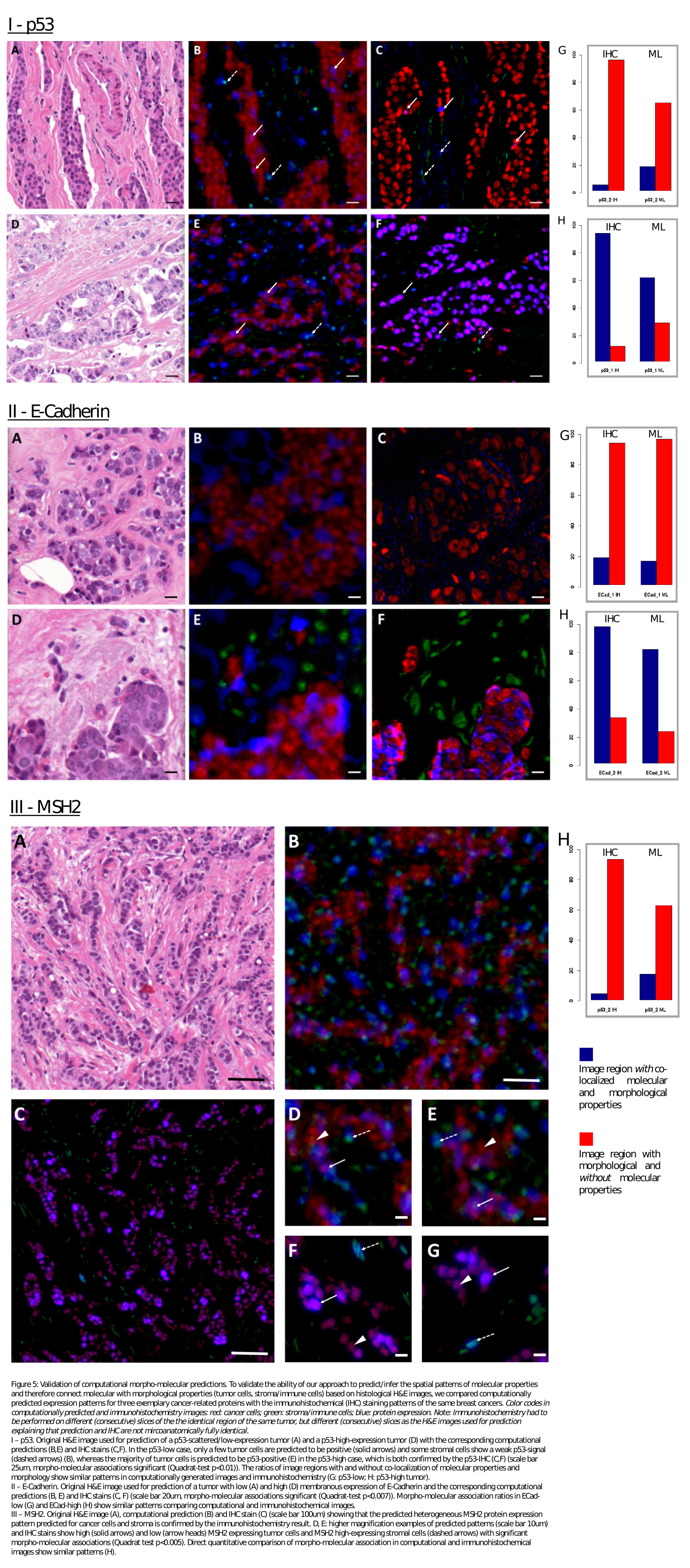}
\captionsetup{labelformat=empty}
\caption{\label{fig:fig5mainms} }
\end{center}
\end{figure}

Similar to the ability of our approach to deduce the localization of cells and tissue features from the initially non-spatial classification result, we are able to deduce spatial probability maps from the molecular feature prediction that indicate the contribution of the respective spatial regions to the classification result (Fig. \ref{fig:fig1mainms},\ref{fig:fig4mainms},\ref{fig:fig5mainms}). The combination of histological and molecular feature prediction can then be used to spatially relate molecular with morphological feature predictions. Although spatial correlation of the molecular prediction with, e.g. cancer cells, stroma or tumor infiltrating lymphocytes does not imply any causal link, as it is statistical in nature, this information may be useful to generate hypotheses on the relevance of certain components of the tumor microenvironment for the presence of certain molecular tumor profile features, which may in the future lead to better causal understanding. Our technique introduces a novel means of computationally generating images of molecular markers that we will denote as "computational fluorescence microscopy", because the probability scores that are assigned to certain image regions can be thought of as equivalent to the intensity of a fluorescent dye indicating the presence of certain molecules stained by fluorescent dyes. However, it is important to emphasize that the computational prediction does not necessarily indicate where the molecular property is localized itself, but that the machine learning approach identifies the spatial / morphological cell and tissues features that are statistically associated with or indicative of the probability of its presence or expression. In addition to predicting the (quantitative) presence of the molecular feature our approach may therefore propose associations between molecular feature and certain microscopic/microanatomic features of the tumor. We computed molecular “heatmaps” connecting molecular and spatial microanatomic information for all 190 proteins and a subset of 409 CNVs from genes known to be genetically altered in cancer. We observed statistically significant spatial correlations for 131 molecular features with cancer cells, TiLs or stroma among data on 190 proteins and 409 CNVs (Fig. \ref{fig:fig4mainms}). The spatio-molecular correlation analysis (Fig. \ref{fig:fig4mainms}A) shows that molecular features – in the majority of the cases – have distinct association with either cancer cells, TiLs or stroma and that also within each of the three morphological groups, molecular markers show significantly different correlation patterns. Average spatio-molecular correlations range from $r=-0.30$ to $r=+0.27$ (mean correlations per molecular feature, $p<0.01$, $\alpha=0.05$), which is considerable given the close proximity (and partially overlapping probability maps) of cancer cells, TiLs and stroma. Such spatio-molecular correlation analysis allows for the definition of candidate lists of molecular features available through the training data being associated with morphological features. Here, top-ranked morpho-molecular associations are positive correlations for TiLs with expression of B-Raf, GSK3-alpha-beta, Cyclin-B1, PI3K-p85 and copy number variations for RBM15. ERBB2 amplification and protein expression, MDM2 and TSC2 copy number variations, for instance, appear to be positively spatially associated significantly with the stromal component of the tumor microenvironment. Top ranked positive spatial correlation with cancer cells are predicted for S6, Chk2 and cMET as well as BRCA2 and TRIM33 amplification. Similarly, top ranked negative associations with TiLs are predicted for Caveolin-1, Collagen\_VI, CyclinD1 expression, whereas S6 as well as BRCA2, TRIM33 and RBM15 copy number gains are the top negatively associated molecular features with respect to stroma. A clustering analysis also shows that correlation patterns of molecular features with carcinoma cells and TiLs are similar in comparison with stroma, offering support for currently intense discussion on the importance of TiLs. 

\subsection{Experimental validation of morpho-molecular predictions}

To test if predictions are correct not only in terms of abundance/expression (as measured by the balanced accuracy), but also with respect to their spatial distribution and association with morphological tissue properties in exemplary cases, we stained nine tumors (triplicates for each molecular marker) randomly selected from routine diagnostics for E-Cadherin, p53 and MSH2 by immunohistochemistry (see supplement). Then, we performed molecular predictions using our machine learning approach for E-Cadherin, p53 and MSH2 based on Hematoxylin-Eosin stained consecutive tissue sections in addition to the identification of cancer cells, stromal cells and lymphocytes. We combined these computationally and experimentally generated images into pseudo-fluorescence images.
	To compare the association of molecular and morphological properties between the computationally generated images and the immunohistochemical stains, the images cannot be overlaid directly because morphological stains used for prediction and immunohistochemistry for validation are generated from sequential sections of the tissue blocks which show similar structures but significant morphological variation perpendicular to the cutting plane where deviations already on the scale of microns impair a pixel-wise statistical comparison (see online methods and supplement: Fig. \ref{fig:tumorstructure}). We therefore used the Quadrat test\cite{gatrell1996} (see online methods and supplement) to demonstrate significance of spatial morpho-molecular associations in both computational and experimental images (Quadrat test p-values: p53 low $p_{ML}=2.8e-07$, $p_{IHC}=8.5e-03$, p53 high $p_{ML}=2.2e-16$, $p_{IHC}=2.2e-16$; E-Cadherin low $p_{ML}=1.3e-03$, $p_{IHC}=2.4e-03$; E-Cadherin high $p_{ML}=6.3e-03$, $p_{IHC}=1.4e-11$; MSH2 $p_{ML}=4.5e-03$, $p_{IHC}=4.7e-06$) and then computed the co-localization ratios rcl of image regions where molecular prediction/stain and morphological properties co-localize and where they are unassociated which provides a quantitative similarity measure ([molec. feature, coloc. ratio IHC, coloc. ratio ML] [p53low, 8.5, 2.2], [p53high, 0.04, 0.27], [MSH2, 2.2, 1.3], [ECadlow, 0.19, 0.16], [ECadhigh, 3.0, 3.6]). Apart from the variations due to consecutive sections, differences such that IHC ratios are higher than ML ratios for p53 and MSH2 can also be partly explained by the fact that IHC precisely stain cell nuclei whereas predictions despite their significant association between molecular marker and cancer cells are blurred resulting in lower ratios. However, the computational predictions reflect the overall patterns of the experimental validations which is obvious when comparing pairs of ratios across markers (Fig. \ref{fig:fig5mainms}). In addition to this experimental validation we generated synthetic histological images composed of different shapes and corresponding molecular properties, for which we then performed the morpho-molecular predictions and generated computational immunohistochemistry images. These tests further confirmed the non-random, biologically meaningful character of the spatial predictions (see methods and supplement for details).

\section{Discussion}

From a methodological perspective our results show that the machine learning-based method we propose can be flexibly used to identify and predict morphological as well as molecular tissue properties or even patient survival (supplement: Fig. \ref{fig:survhm}+\ref{fig:kmplot}) from various microscopic imaging sources in breast cancer. While it poses a challenge to provide training data covering the broad variability particularly in cancer tissues (for which we created a publicly available pathologist-annotated cancer and normal cell image data base), and while the method is limited by technical variability in tissue sectioning and staining, our analyses show a good performance with reasonable amounts of training data and robustness to technical variation (see supplement: Fig. \ref{fig:suppl_spatfig}). In terms of cancer biology, our results demonstrate a close link between morphological properties and molecular properties related to cancer. While the classical grading system used in diagnostics is based on the notion that increasing genomic instability is reflected by a morphological disarray (a concept that we recently confirmed by showing that this correlates with increasing mutational frequencies\cite{doi:10.1002/cjp2.25}), our present study demonstrates that individual molecular features including gene expression, gene methylation, copy number variations, somatic mutations and protein expression and phosphorylation that are related to cancer can be predicted – to a certain degree – from histological data. Among the different molecular features we predicted, DNA methylation and gene expression (RNASEQ) data yielded, on average, better prediction scores than mutations (simple somatic mutations and copy number variations). It may on one side be speculated that this is because the former features are downstream of DNA and therefore also closer to morphological phenotypes. However, on the other side, we did not observe better predictability for proteins, although this may be related to the comparatively higher noise associated with the method used for protein measurements. 
	We validate the molecular predictions first by out-of sample predictions on their presence or abundance and by showing that many of the markers with relatively high scores are known to be cancer related by gene ontology enrichment. Furthermore, we performed a statistical analysis based on Hoeffdings inequality corrected for multiple testing for molecular predictions. Associations with GO categories were validated through a control experiment by repeating the analysis with randomly permuted labels (Fig. \ref{fig:fig3mainms}B). Unlike the analysis using the correct labels (Fig. \ref{fig:fig3mainms}A), the control experiment yielded, as expected, no correlations with the gene ontology information, demonstrating a clear biological significance of our ML-based molecular prediction approach (Fig. \ref{fig:fig3mainms}A). Consequently, in addition to using our approach to support molecular “wet-lab” analyses by “computational molecular profiling”, the method can also be used to help score the cancer-relatedness of novel or so far unexplored molecular features.  
Combining the ability to identify cell and tissue features in microscopic data with the capability to not only predict molecular features but statistically infer which image regions contribute most to the molecular predictions, our approach can help reveal relations between non-spatial molecular and spatial microscopic information. As a validation of the molecular heatmaps ("computational fluorescence images"), we show the concordance between predicted location and intensity with immunohistochemical staining patterns for different molecular properties and use the Quadrat method\cite{gatrell1996} to establish significance of morpho-molecular associations. While this demonstrates that biological meaning can be inferred by our machine learning approach, it has to be noted, that predictions can also be meaningful if predicted and physical location show discrepancies (for instance, a protein may be expressed in the tumor cells, but induced by and therefore spatially associated with certain tumor stroma properties). As additional validation, we generate synthetic data, in which molecular properties are assigned to different geometric shapes corresponding to morphological tissue components. Performing the same morpho-molecular predictions for simulated images containing different random compositions of these shapes demonstrates that molecular properties are spatially co-localized with the corresponding morphological shapes in a non-random fashion. 
	We also evaluated the potential clinical utility of our approach by testing its capability to predict hormone receptor status, grading and overall survival from morphology. While the observed balanced accuracies are not yet sufficient for the high demands of clinical routine application, we show that it is possible to automatically identify subsets of cases with classification accuracies of up to 98\% based on the SVM scores. The approach may therefore be used to apply molecular profiling in a more targeted manner and to complement immunohistochemistry and support visual diagnostic evaluation. With respect to the survival analysis, we computed heatmaps to identify prognostically relevant image regions and observed that tumor-infiltrating lymphocytes (TiLs) are associated with shortened survival in luminal breast cancer (see supplement: Fig. \ref{fig:survhm}), which is in line with initial evidence we reported recently on the different prognostic relevance of TiLs in luminal vs. basal breast cancer\cite{Denkert2018TumourinfiltratingLA}. These observations should be considered carefully and as indicators of a potential clinical utility of the morpho-molecular predictions, clearly they will certainly require further validation in clinical trials.
	To conclude, not proving any causality by itself, the machine learning approach we present can be used to help interpret experimental data and generate hypotheses on possible mechanistic links with implications for basic science and clinical medicine. It will become increasingly powerful with the growing amount of molecular profiling data becoming available and can therefore reconcile non-spatially resolved high-dimensional molecular data with spatial (microenvironmental) information on cell and tissue features not only in the context of cancer but potentially any complex multi-cellular biological system. 

\section*{Contributions}

Conceptualization: AB, MB, MH, KRM, FK. Methodology: AB, MB, MH, SW, DH, KRM, FK. Software: AB, MB, MH, SW, DH, KRM, FK. Validation: AB, MB, MH, KRM, FK. Formal Analysis: AB, MB, MH, FK. Investigation: AB, MB, MH, CD, KRM, FK. Resources: AB, MK, MI, MD, CD, KRM, FK. Data Curation: AB, MB, DH, LP, FK. Writing – Original Draft: AB, KRM, FK. Writing – Review \& Editing AB, MB, MH, SW, DH, AS, LP, JB, DT, MK, MI, MD, KH, AH, CD, KRM, FK. Visualization AB, MB, MH, FK. Supervision: KRM, FK. Funding: AB, MI, KRM, FK.

\section*{Acknowledgements}

This work was funded by the Charité Institute of Pathology, Berlin, the Technical University of Berlin), the Human Frontier Science Program (HFSP) Young Investigator Grant (M.K., M.I. and F.K.) and the Einstein Foundation Berlin (F.K.) and partly by the German Research Foundation to A.C.H (DFG SFB-TR84, B6, Z1a) and the German Consortium for Translational Cancer Research (DKTK). K.-R.M. thanks DFG, BMBF (Berlin Big Data Center) and the Institute for Information \& Communications Technology Promotion (IITP) grant funded by the Korea government (no. 2017-0-00451). A.B. thanks the SUTD for the SUTD startup grant.

\bigskip

\phantomsection
\addcontentsline{toc}{section}{References}
\bibliographystyle{apalike}
\bibliography{techsupplement}

\section{Supplemental Material: General description of BoW features}
\label{sec:bow}
Here we describe generically the Bag-of-Words (BoW) features, which are the basis for molecular profile prediction, molecular property heatmapping, cancer heatmapping and lymphocyte heatmapping. BoW features in computer vision \cite{Csurka04} originate from document analysis research, which can be traced back to the vector space model of \cite{Salton:1975:VSM:361219.361220}. They are highly successful in concept recognition as they have been a key ingredient in top-ranked submissions to international competitions in that field constantly over the last years \cite{pascal-voc-2008,pascal-voc-2009,NowakD09},\newline\cite{NowakH10}. This part serves as an introduction of BoW features to readers who are not familiar with this field. Specific details on parameters of the BoW features used in the experiments are given further below. 

\subsection{BoW feature computation pipeline}

Prior to analysis, images are split into tiles as we computed a set of BoW features for single tiles. A feature here merely denotes a vector of values used for further prediction tasks. The computation of BoW features consists of four steps described below. One step is required only once during the training phase, namely the generation of the vocabulary of visual words (sometimes called prototypes).

\ben{
\item \textbf{Select regions for local feature computation} \\
First, regions are selected from which local features are computed. One can use for example regions over points centered on a grid. This is a common case in image classification on ordinary multimedia images and is used in this paper. In contrast to single-cell- or segmentation-based pipelines these regions do not need to coincide well with cells or cell boundaries. 

\item \textbf{Extraction of local features over regions}\\
In the next stage local features are computed over the regions obtained in the last step. Technically, a local feature is a vector resulting from an algorithm applied to a region in the image tile. Local features can be the well known SIFT descriptor \cite{Lowe04} (without the method to generate keypoints) or for example the mean and variance of pixel intensity over a region. Practically, we compute multiple types of descriptors from the RGB color channels in an image. Details of how to compute local features will be given further below for each task.

\item \textbf{Training phase only: Create a set of visual words}\\
During training stage the local features from a subset of training images are used to compute a set of visual words for example by clustering methods. A visual word is formally a point in the space of the local features. This enables to compute distances in the space of local features between the features and the set of visual words. These distances are the key to compute a BoW feature. It is important that a visual vocabulary is kept fixed once it has been created, as the BoW feature is based on distances between local features and the visual words. The visual words are used for computing the global BoW feature from training as well as test images.

\item \textbf{Compute the Global BoW Feature} \\
Finally, one global BoW feature is computed for each tile. A BoW feature is a histogram, which gets normalized. Each bin of the histogram represents one visual word from the set of visual words. Thus, the dimensionality of a BoW feature is the number of visual words in the vocabulary. The coarse idea is to compute distances to the visual words for each local feature. With this notion of distance between the local features as an additional information, more elaborate mapping schemes such as soft codebook weighting \cite{DBLP:conf/eccv/GemertGVS08} are applicable. Then one can aggregate all local feature mappings into one global feature, usually by summing them. This aggregation step allows to use images of varying sizes, and varying numbers of local features, and to arrive at a BoW feature of fixed dimensionality.
}

\noindent This procedure is repeated for different kinds of local feature extraction algorithms (like SIFT or mean image intensity of a region) and different parameters (like the size of the region used to compute a local feature, or color channels in the image) resulting in a set of different BoW features for each image tile.

\subsection{Training and classification with BoW Features}

Here one has to discriminate between an initial training stage and a testing (out of sample) stage. The computed set of BoW features is used to train a classifier during training stage. A classifier $f$ is a mapping which takes a feature $x$ as input (here the set of BoW features) and computes a real-valued score, which is used for prediction of the output: $f: x \longmapsto f( x) \in \RR$. For supervised learning, this requires labels for all image tiles in the training dataset.

The training stage can be described as the (usually supervised) selection of a classifier mapping $f$ from a set of hypotheses $\mc{F}$ using training data yielding a set consisting of pairs of BoW features $x_i$ and their corresponding labels $y_i$:
\be
\label{bowformula2}
\{(x_1,y_1),\ldots, (x_n,y_n) \} \longmapsto f \in \mc{F}
\ee
Typically, one BoW feature corresponding to one kind of local feature is not sufficient to capture enough information to achieving good classification performance, instead multiple types of BoW features corresponding to multiple local features are required. In practice we compute a set consisting of different global BoW features from each image tile. Each BoW feature is defined by a specific choice of parameters of the BoW approach, for example, the local feature, the number of visual words or the support area size of the local feature regions. 

During the testing stage a set of BoW features $x$ is computed per subimage on which the trained classifier f is then applied. The output $f(x)$ is a real-valued measure for the presence of the predicted parameter, e.g. cancerous structures. Thresholding of the score can be used to obtain binary predictions.

\subsection{General remarks}
\label{ssec:bowremarks}

A BoW-based approach does not require exact labels for cells or regions within in the image tile. It is sufficient to provide labels in $\{-1,+1\}$ for the whole image tile indicating whether the image tile does or does not contain the relevant structure. This requires less labeling effort and is adapted to our prediction problem, which deals with predicting the presence of cancerous structures for an image tile. 

The local feature extraction is typically done such that local feature regions are overlapping. This reflects experimental evidence that sparse sampling of keypoints is detrimental for BoW model performance \cite{DBLP:conf/eccv/NowakJT06}. It accounts for the potentially lower information content of a generic local feature compared to specialized morphological features which have been constructed using prior knowledge. The BoW feature replaces deterministic decisions for segmentation masks by statistic sampling of local features (cf. also for a description of BoW in multimedia image ranking problems \cite{DBLP:journals/cviu/BinderSMK13}).

\subsection{Local features}
\label{ssec:basefeat}

Here we describe our set of base BoW features. A part of this set of BoW features is used for all image stain (e.g. H\&E) modalities. For different stains, different combinations of these base features with different methods of region selectionfor local feature computation were used, further, different sets of color channels for computing the local features for these base features were applied.

\subsubsection{Local features: SIFT}
\label{ssec:lf0}
We used the SIFT \cite{Lowe04} feature implementation from version 0.9.9 of VLFeat \newline\cite{vedaldi08vlfeat}. In particular we applied the SIFT descriptors with rotation invariance but with densely sampled keypoints. 

\subsubsection{Local features: Gradient Norm Quantiles}
\label{ssec:lf1}
\begin{figure}
\centering
  \includegraphics[width=0.18\textwidth]{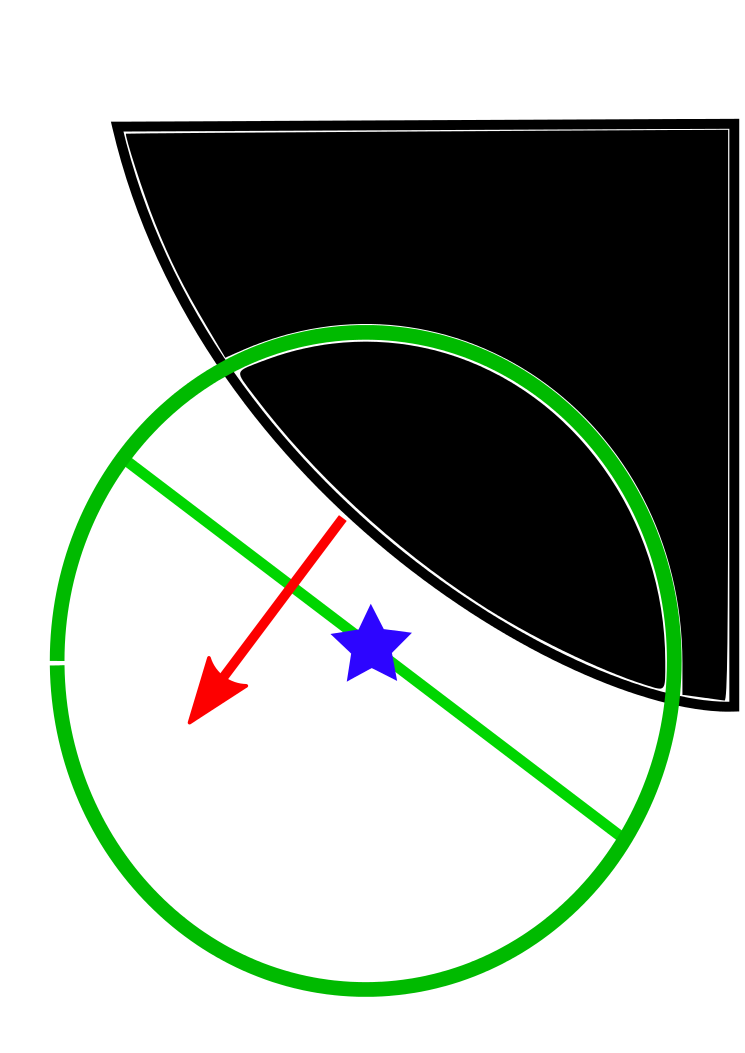}
\caption{\label{fig:lfgeom} Local feature geometry used for computations of quantile-based local features. A circular region (green circle) centered on the keypoint, given by the blue star, is divided in two halves. The red arrow depicts the dominant direction of the gradient norms within the green circle. The dividing line for the two halves is orthogonal to the dominant direction of gradients computed within the circular region.}
\end{figure}

The Gradient Norm Quantiles (gnq) feature is computed over a circular region of radius equal to six times the feature scale in consistency with conventions used from SIFT features. The circular region is divided into two half circles of equal size. The dividing line is chosen orthogonal to the dominant direction of gradients computed within the circular region. Figure \ref{fig:lfgeom} depicts this construction. For all local features we used the computation of dominant orientations implemented in VLFeat \cite{vedaldi08vlfeat}.

In each half circle a 9-dimensional vector is computed. Its $i$-th dimension corresponds to the value of the quantile estimator for the quantile $i/10, 1 \le i \le 9$ computed from the gradient norms over all pixels in the half circle. The gradient norms are computed as $\ell_2$-norms from the gradient outputs of a Sobel filter without blurring. Concatenating both 9-dimensional vectors yields an 18-dimensional local feature. The quantile estimator $\widehat{E}(q,V)$ used for quantile $q$ and a sorted set of values $V$ of size $N$ is given in eq.~\eqref{eq:quantest}. In case of multiple color channels, the 18-dimensional vectors are concatenated.
\begin{align}
I & =q(N-1)\\
\widehat{E}(q,V)& = (1- (I -\lfloor I \rfloor ) )V( \lfloor I \rfloor +1 ) + (I -\lfloor I \rfloor )V( \lfloor I \rfloor +2) \label{eq:quantest}
\end{align}

\subsubsection{Local features: Intensity Quantiles}
\label{ssec:lf2}

The Intensity Quantiles (ciq) feature is also computed over a circular region of radius equal to six times the feature scale in consistency with conventions used from SIFT features. Again the circular region is divided into two half circles of equal size. The dividing line is chosen to be orthogonal to the dominant direction of gradients computed within the circular region. Figure \ref{fig:lfgeom} depicts this construction. For all local features we used the computation of dominant orientations as in VLFeat \cite{vedaldi08vlfeat}.

In each half circle a 9-dimensional vector is computed. Its $i$-th dimension corresponds to the value of the quantile estimator for the quantile $i/10, 1 \le i \le 9$ computed from the image intensities over all pixels in the half circle. The pixel intensities are taken directly from the pixels. Concatenating both 9-dimensional vectors yields an 18-dimensional local feature. The quantile estimator $\widehat{E}(q,V)$ used for quantile $q$ and a sorted set of values $V$ of size $N$ is given in eq.~\eqref{eq:quantest}. In case of multiple color channels, the 18-dimensional vectors are concatenated.

\subsubsection{Local features: Concatenations}
\label{ssec:lf3}

There are two possible approaches: high level fusion and low level fusion. High level fusion denotes the following approach: One uses multiple local feature types, computes one BoW feature from each local feature type, and then fuses the BoW features on the level of a kernel or support vector machine. In this approach we obtain one BoW feature from each single type local feature. Here, type refers to one method to compute a local feature. But also types are different if parameters of the local feature are different, i.e. gradient norm quantiles with different feature radii or computed over different sets of color channels would constitute two different types.

Low-level fusion denotes the usage of composite local features which are concatenations of local feature types, and then a single BoW feature from this concatenated local feature is computed. The features SIFT+gnq, SIFT+ciq and
SIFT+ciq+gnq in Table \ref{tab:baseconfs} refer to such concatenations of local features. In this case, we used a weighted $\ell_2$-metric for clustering: at first we divided each dimension of local features by its standard deviation, which has been obtained from a large set of local descriptors, then we upscaled the local feature dimensions inversely proportional to the dimensionality of the compound descriptor. E.g. for a concatenation of SIFT and ciq, after normalization, the dimensions corresponding to SIFT (128-dimensional) got multiplied by 1/128, the dimensions corresponding to ciq got multiplied by 1/18.

\subsubsection{Color channels}
\label{ssec:colchannelsgeneral}

For improvement of performance, local features are often computed over a set of different color channels and concatenated \cite{DBLP:journals/pami/SandeGS10}. This allows to incorporate color information and correlations between various color channels. Examples for such sets of color channels are the basic set of red, green, and blue (RGB), the set (OPP) composed of the three channels grey \eqref{eq2:greycolorchannel}, opponent color 1 \eqref{eq2:o1colorchannel} and opponent color 2 \eqref{eq2:o2colorchannel}. They are given in eq. \eqref{eq2:greycolorchannel},\eqref{eq2:o1colorchannel},\eqref{eq2:o2colorchannel} as functions of RGB-values $(r,g,b)$ lying in $[0,1]$.
\begin{align}
gr(r,g,b)&= (r+g+b)/3 \label{eq2:greycolorchannel}\\
o1(r,g,b)&=(r-g+1)/2 \label{eq2:o1colorchannel}\\
o2(r,g,b)&=(r+g-2b+2)/4 \label{eq2:o2colorchannel}
\end{align}

\begin{table}
\centering

\caption{\label{tab:baseconfs} Base local feature types when used for a single color channel}
\begin{tabular}{cccc}
\hline
type & scale & feature radius & local feature dim. \\
 & & & per color channel  \\ \hline
gradient norm quantiles (gnq) & 1.5/2.0/2.5 & 9/12/15 & 18  \\
color intensity quantiles (ciq) & 1.5/2.0/2.5 & 9/12/15 & 18  \\
SIFT & 1.5/2.0/2.5 & 9/12/15 & 128  \\
SIFT+gnq & 1.5/2.0/2.5 & 9/12/15 & 146  \\
SIFT+ciq & 1.5/2.0/2.5 & 9/12/15 &146\\
SIFT+ciq+gnq & 1.5/2.0/2.5 & 9/12/15 & 164 \\
\hline
\end{tabular}
\end{table}

\subsection{Training phase: Visual word generation}
\label{ssec:vwgen}

We used k-means to compute visual codebooks. The local features used are a subset of the local features extracted from the training data set.

\subsection{BoW mapping function}
\label{ssec:bowmap}

We used rank-based BoW mapping with a cut-off constant $k=4$. Let $RK_d(l)$ be the rank of the distances between the local feature $l$ and the visual word corresponding to BoW dimension $d$, sorted in increasing order. Then the BoW mapping $m_d$ for dimension $d$ is defined as:
\be
m_d(l)= \begin{cases} 
2^{-RK_d(l)} & \mbox{ if } RK_d(l) \le 4 \\
0	& \mbox{ else.}
\end{cases}
\ee
\subsection{BoW normalization}
\label{ssec:bownorm}
We used $\ell_1$-normalization.
\be
x \longmapsto \frac{x}{\|x\|_1}
\ee

\section{Supplemental Material: Molecular profile prediction}
\label{sec:mol}

The goal in this task is two-fold. First, we apply our machine learning approach to predict molecular features from histomorphological images. Second, we use our approach to identify those image regions (down to single pixels) that contribute most/least to the molecular predictions. For each pixel a score is determined that describes its relevance for the classification output. These scores are then mapped onto colors in order to yield the computationally generated ``fluorescence'' images  shown in the main paper. The pixel-wise evidence is computed using layer-wise relevance propagation (LRP) \cite{bach2015pixel, samek2016interpreting}. LRP is a principled approach that decomposes the classifier prediction score, computes relevance scores for intermediate feature representations and projects them back through the structures of the classifier onto the input pixels. This approach has been evaluated qualitatively and quantitatively \cite{samek2016evaluating,DBLP:journals/pr/MontavonLBSM17} on BoW features.

\subsection{Classifier: Labels used for molecular profile prediction}

We perform predictions for all major molecular data modalities: Somatic mutations (SOM), copy number variations (CNV), gene expression (RNASEQ), DNA methylation data (METH) and protein profiles from reverse-phase protein arrays (PROT). All modalities (except SOM) consist of real valued scores obtained from measurements for each protein or gene per patient, whereas SOM data contains binary classification labels for each gene. 

The available samples sizes at the time of analysis from TCGA were 554 for CNV, 563 for RNASEQ, 400 for METH and 565 for PROT. SOM was a special case because only 23 genes were mutated in more than 10 samples. Due to this, we restricted our analysis to this subset.

The prediction was done for each protein/gene separately. Considering the anticipated amount of noise typical for a biological prediction task and the small sample size, we decided to use a high versus low classification approach instead of regression. By thresholding the data of the patient population for the selected protein/gene, we obtained binary labels for the molecular profiling data. Values above the threshold were labeled $+1$, values below $-1$. We used as thresholds the set of all 9 equispaced quantiles between $\{0.1 \cdot i, 1 \le i \le 9\}$ of the value of the whole population for the protein/gene measurement. This approach ensured that the smallest class in the classification problem had at least 40 samples. It also accounts for the fact that for the majority of considered genes/proteins there is no known cut-off which  separates the patient population in groups with distinct biological behavior.

\subsection{Classifier: Parameters of the BoW feature used for molecular profile prediction}

In this paragraph we will give details on the features used and their parameters. For a technical description of the Bag-of-Words (BoW) features in all generality please refer to Section \ref{sec:bow}. Features were computed as follows: At first an image was divided into tiles of size $201\times 201$. The tiling was created with a stride of $201/3=67$ moving along a regular grid. We computed one BoW feature per tile. 

We considered not only cell cores but also stroma for the prediction task. This required us to be robust against stain variations present in the H$\&$E stain. We therefore only used SIFT features here as local feature in order to be maximally invariant to stain variations and in order to mainly capture texture information. For this reason, BoW features over color quantiles and quantiles of norms of gradients were omitted. We used SIFT as local feature with scale $2.0$ corresponding to a feature radius of $12$ pixels. The local feature radius was chosen to be slightly larger than the size of cancer nuclei observed. Choosing local feature radii slightly smaller and larger than nuclei showed good results for cancer evidence detection tasks. The local feature was computed over red and blue color channels resulting in a 256-dimensional feature and sampled on a grid of stride 3. At training time a visual vocabulary of $510$ words was computed by k-means clustering. The number of visual words was chosen based on prior knowledge and prior evidence from experience with a different task, namely cancer presence prediction in image patches. 

There is no strong sensitivity in the visual vocabulary size for the task of cancer presence prediction in image patches. The relatively small sample size for the molecular profile prediction, in the order of hundreds of samples, as well as the fact that we had thousands of different proteins and genes to predict implies a certain risk of overfitting. Therefore, although optimal vocabulary size might be individually different, we refrained from any optimization of the vocabulary size for all those proteins and genes. 

The final BoW feature was computing by mapping the SIFT features by rank-weighted soft mapping \cite{DBLP:journals/cviu/BinderSMK13}, summing the resulting mapping vectors and normalizing them to unit $\ell_1$-norm, resulting in a 510 dimensional BoW feature. The BoW features over all tiles were averaged into one global feature per image. This global BoW feature was used to make a prediction.

\subsection{Classifier: Kernel computation for molecular profile prediction}

We used a histogram intersection kernel (HIK). The choice of histogram intersection kernel for this setup against the alternative of the $\chi^2$-kernel was motivated by details of the heatmapping procedure. The usage of the $\chi^2$-kernel would have required a larger set of images which would be used only to fit an appropriate root of the Taylor expansion. Given the small sample sizes of at most $565$ patients, we decided not to split this dataset further. This problem does not arise for the cancer evidence detection task, where a validation set of over $4000$ patches was available. The HIK is defined by: 
\eqf{
K(x,z)= \sum_d \min(x_d,z_d)  \ .
}
The normalization constant of the kernel is chosen as the standard deviation in Hilbert space, that is:
\eqf{
c &= \frac{1}{n}\sum_{i=1}^n K(x_i,x_i) - \frac{1}{n^2}\sum_{i=1}^n \sum_{j=1}^n K(x_i,x_j) \\
K & \mapsto \frac{K}{c} 
}
The constant was computed over the kernel from the training samples. This ensures that the optimal SVM regularization constant is in practice close to $1$.

 \subsection{Classifier: Learning algorithm for molecular profile prediction}

A dual support vector machine (SVM) from the shogun toolbox \cite{SonRaeHenWidBehZieBonBinGehFra10new} was used to predict the data from the averaged BoW feature. The only tunable parameter is the regularization constant of the SVM.

\subsection{Classifier: Experimental setup for molecular profile prediction}

The performance reported is the mean of true positive and true negative rate, which we termed balanced accuracy in the manuscript. In particular, the baseline under this measure obtained by a random or a constant prediction is always 50\%, even if we classify 10\% of the dataset as belonging to one class against the remaining 90\%. 
We do not report accuracy because the accuracy of a constant prediction for a dataset with 90\% positive labels is 90\%, suggestive of an excellent performance, whereas the balanced accuracy for the same predictor will be only 50\%. Unlike the equal error rate (EER) the balanced accuracy does not involve a posteriori optimization over thresholds which cannot be applied in prediction tasks at deployment time (due to lack of ground truth at deployment time), but uses the classification thresholds from the predictor as they are. We consider the balanced accuracy to be more suitable for comparison of results with small and variable class-ratios.

Given the small sample size, the experiment was run for each protein and threshold quantile with 10-fold outer cross-validation and 9-fold inner cross-validation. The splits were randomized with the constraint, that for a k\% quantile as threshold, each split contained approximately k\% of the smaller values. For each step of the outer cross-validation, the best regularization constant was chosen by the error rate measured on the 9-fold inner cross-validation - which was performed on the training set of the same outer cross-validation step. Thus, the test set of each step of the outer cross-validation was never used to optimize parameters for the same step. It was only used to measure accuracies after the regularization constant has been determined. The performance was reported on the test set of each outer cross-validation step. The reported result is an average over the outer cross-validation steps. The result is shown for the best-performing quantile. Given the small sample size, choosing a single test set might result in increased variance. Regularization constants for the cross-validation were taken between from $10^{p/2}$, $p=\{-2,...,+2\}$.

\subsection{Heatmapping: Molecular profile prediction}

The heatmapping was done as described for BoW features in \cite{bach2015pixel}. Scores on kernel level were redistributed to the BoW features over all overlapping tiles, and then, for each tile, backprojected onto local features and pixels.

The heatmap for the protein expression levels indicates localized evidence for values above the population quantile threshold (ascending from green to yellow to red) versus evidence for values below the same threshold (ascending from green to blue). Neutral regions without any particular evidence are colored in green.

For better understanding of the following formulas we introduce notation. The input $x$ for classification of an image is a concatenation of BoW features where $x_{d,u}$ is the $d$-th dimension of the $u$-th BoW feature in the concatenation. A BoW feature is computed as a sum of mappings $m$ of local features onto visual words. $m_d^{(u)}$ is the mapping onto the $d$-th dimension of the $u$-th BoW feature.
\be
x_{d,u}= \sum_{i=1}^N m_d^{(u)}(l_i)
\ee
Each local feature $l$ is used only for one BoW feature where $T(l)$ is the index of that BoW feature.
The SVM prediction for an image and multiple kernels $k_u$ can be written as:

\be
f(x)=b+\sum_{i=1}^S\alpha_iy_i \sum_{u=1}^K \beta_u k_u(x_i,x)
\ee
Before we assign a score to each pixel in an image, we need to define a local feature score $R$ for a local feature. We seek a set of local feature scores $R(l)$ for all local features such that its sum approximates the SVM prediction.

\be
f(x)\approx \sum_l R(l) \label{eq:lscoreapprox}
\ee
This is achieved in two steps. At first we compute a relevance score $r_{d,u}(x)$ for the dimension $d$ of the $u$-th BoW feature. In the second step we compute a local feature score $R(l)$ using the relevance score $r_{d,u}(x)$ in accordance to equation \ref{eq:lscoreapprox}.

For the HIK kernel the relevance score for a BoW dimension can be computed ad-hoc as has been observed already in \cite{DBLP:journals/ijcv/UijlingsSS12}.

\eqf{
k_u(x_i,x) &= \sum_{d=1}^{D_u} \min(x_i^{(d)},x^{(d)}) \label{eq:hikstart} \\
\Rightarrow f(x) &=b+\sum_{i=1}^S\alpha_iy_i \sum_{u=1}^K \beta_u \sum_{d=1}^{D_u} \min(x_i^{(d)},x^{(d)}) \\
& = \sum_{u=1}^K \sum_{d=1}^{D_u} \frac{b \beta_u}{\sum_{u'} \beta_{u'} D_u}+ \beta_u\sum_{i=1}^S\alpha_iy_i \min(x_i^{(d)},x^{(d)})\\
f(x) & = \sum_{u=1}^K \sum_{d=1}^{D_u}r_{d,u}(x)\\
\Rightarrow r_{d,u}(x) & = \frac{b \beta_u}{\sum_{u'} \beta_{u'} D_u}+ \beta_u\sum_{i=1}^S\alpha_iy_i \min(x_i^{(d)},x^{(d)}) \label{eq:hikend}
}
For the special case of molecular profile data prediction, the BoW feature is an average of BoW features of single overlapping tiles. This is done merely for the sake of being able to precompute features in a distributed manner.
\eqf{
x^{(d)}  = \frac{1}{T}\sum_{t=1}^T x^{(d,t)}
}
The relevance score obtained for the dimension $d$ of the composite BoW feature $x$ needs to be backprojected onto the BoW features for each tile. This is done by naturally following the linear structure:
\eqf{
r_{d,u}(x)  = \frac{1}{T}\sum_{t=1}^T r_{d,u}^{(t)}(x)
}
$r_{d,u}^{(t)}(x)$ is the relevance score for BoW feature dimension $d$ of BoW feature $u$ in image tile $t$.
In the following we will write  $r_{d,u}(x)$ as a slight abuse of notation. The relevance scores $r_{d,u}(x)$ for single dimensions of BoW features permit us now to define a score for local features $R(l)$ via
\be
R(l):= \sum_{d \not\in Z_{T(l)}(x)  } r_{d,T(l)}(x) \frac{m_d^{(T(l))}(l)}{\sum_{i \mid T(i)=T(l) } m_d^{(T(l))}(i)} + \sum_{d \in Z_{T(l)}(x)  }  r_{d,T(l)}(x)\frac{ 1 }{\sum_{i  \mid T(i)=T(l) } 1} \label{eq:locfeatformula2}
\ee
where $Z_k$ is the set of dimensions onto which no local feature is mapped:
\be
Z_{k}(x)=\left\{d \mid \sum_{i \mid T(i)=k} m_d^{(k)}(i)=0 \right\}
\ee
Note that we assume $0/0 = 0$. Eq.~\eqref{eq:locfeatformula2} has the property that summing over all local features yields the sum of the BoW feature relevances $r_{d,u}(x)$. This property is the justification for the definition of the relevance of a local feature according to eq.~\eqref{eq:locfeatformula2}.

\begin{align}\sum_l R(l) & = \sum_l \left( \sum_{d \not\in Z_{T(l)}(x)  } r_{d,T(l)}(x) \frac{m_d^{(T(l))}(l)}{\sum_{i \mid T(i)=T(l) } m_d^{(T(l))}(i)} + \sum_{d \in Z_{T(l)}(x)  }  r_{d,T(l)}(x)\frac{ 1 }{\sum_{i  \mid T(i)=T(l) } 1} \right) \nonumber
\\ & = \sum_{u=1}^K \sum_{l \mid T(l)=u} \left( \sum_{ d \not\in Z_{T(l)}(x) } r_{d,T(l)}(x) \frac{m_d^{(T(l))}(l)}{\sum_{i \mid T(i)=T(l) } m_d^{(T(l))}(i)} \right) \nonumber
\\ \ldots & +   \sum_{u=1}^K \sum_{l \mid T(l)=u} \left( \sum_{ d \in Z_{T(l)}(x) }  r_{d,T(l)}(x)\frac{ 1 }{\sum_{i  \mid T(i)=T(l) } 1} \right) \nonumber \\
& = \sum_{u=1}^K \sum_{l \mid T(l)=u} \left( \sum_{ d \not\in Z_{u}(x) } r_{d,u}(x) \frac{m_d^{(u)}(l)}{\sum_{i \mid T(i)=u } m_d^{(u)}(i)} \right) \nonumber
\\ \ldots & +   \sum_{u=1}^K \sum_{l \mid T(l)=u} \left( \sum_{ d \in Z_{u}(x) }  r_{d,u}(x)\frac{ 1 }{\sum_{i  \mid T(i)=u } 1} \right) \nonumber \\
& = \sum_{u=1}^K  \left( \sum_{ d \not\in Z_{u}(x) } r_{d,u}(x) \frac{\sum_{l \mid T(l)=u} m_d^{(u)}(l)}{\sum_{i \mid T(i)=u } m_d^{(u)}(i)} \right) \nonumber
\\ \ldots & +   \sum_{u=1}^K  \left( \sum_{ d \in Z_{u}(x) }  r_{d,u}(x)\frac{ \sum_{l \mid T(l)=u} 1 }{\sum_{i  \mid T(i)=u } 1} \right) \nonumber \\
& = \sum_{u=1}^K \sum_{ d \not\in Z_{u}(x) } r_{d,u}(x) +  \sum_{u=1}^K\sum_{ d \in Z_{u}(x) } r_{d,u}(x)\nonumber \\
& =\sum_{u=1}^K \sum_{ d=1} ^{V_u}  r_{d,u}(x)
 \label{eq:just}
\end{align}
Finally the relevance of a pixel is the sum of the relevances scores over all local features which cover that pixel taken from one  tile.

\be
rel(p)= \sum_{l \mid p \in \mathrm{support(l)}} R(l) \label{eq:unnormpixscores}
\ee
We normalize the scores by the maximal absolute value of all scores $rel(p)$ from eq.~\eqref{eq:unnormpixscores} in order to obtain the heatmap of an image tile:

\be
hm(p)=\frac{rel(p)}{\max_{s \in \mathrm{Tile}} |rel(s)| } \label{eq:normheatmap}
\ee
The final per-pixel scores are obtained by taking the mean from the relevance over all tiles containing this pixel. After that the scores are normalized to lie in the interval $[-1,+1]$. One obtains one score per pixel for every protein/gene. The set of scores for a set of proteins/genes is then mapped onto multiple colors in order to obtain the plots shown in the main paper.

\section{Supplemental Material: Cancer evidence prediction and heatmapping}
\label{sec:can}
The approach here consists of two steps. First, we train a support vector machine (SVM) classifier and use it to output a real-valued score indicating the confidence of cancer in one image patch. In the second step the prediction score per patch is decomposed into a score per pixel using LRP. The scores of all pixels of one patch are then averaged across patches and normalized to lie in the interval $[-1,+1]$. Mapping the scores onto colors yields then a heatmap (i. e. a computationally generated microscopic image) in which cancer cells can be visually identified by colors from orange to red and normal cells by shades of blue. In contrast to the task of Molecular Profile Prediction (Section~\ref{sec:mol}) we use multiple features and a $\chi^2$-Kernel. 

\subsection{Classifier: Labels for cancer evidence prediction}

The label for each tile in the training and evaluation set was binary, $+1$ if it contained at least one cancer cell, and $-1$ otherwise. We did not create any pixel-wise labels.

\subsection{Classifier: Parameters of the BoW features used for cancer evidence prediction}
\label{ssec:bowforcancer}

At first an image was divided into tiles of size $102 \times 102$. The tiling was created with a stride of $102/3=34$ moving along a regular grid. We computed three BoW features per tile. SVM prediction and heatmapping were done for each tile separately. Finally the heatmaps per tile were overlaid over the whole input image.

The local feature was a concatenation of SIFT and gradient norm quantiles. Its structure is described in detail in Section \ref{sec:bow}. Here we provide details on the parameters used for its computation. It is computed and concatenated over red and blue color channels. The local feature was computed over red and blue color channels resulting in a $292$-dimensional feature, sampled on a grid of stride $3$. We used scales $1.5$, $2.0$ and $2.5$, corresponding to local feature radii of $9$, $12$ and $15$.

At training time a visual vocabulary of $510$ words was computed by k-means clustering. We did not finetune the vocabulary sizes experimentally. From our experience with BoW features for histopathological tasks, typically a vocabulary size of $384$ to $600$ is a good choice for the SIFT local feature. 
The local feature used here is a concatenation of SIFT and quantile based local features, with consequently slightly higher dimensional local feature space compared to a SIFT feature (for one color channel 128+18 as compared to 128 for SIFT only). Thus we have chosen a vocabulary size similar to the size used for SIFT for sufficient coverage of the local feature space. The final BoW feature was computing by mapping the local features by rank-weighted soft mapping, summing the resulting mapping vectors and normalizing them to unit $\ell_1$-norm, resulting in a $510$ dimensional BoW feature.

\subsection{Classifier: Kernel for cancer evidence prediction}

We used a $\chi^2$-kernel: 
\eqf{
K(x,z)= \exp(-\sigma \sum_{d \mid x_d+z_d > 0}  \frac{(x_d-z_d)^2}{x_d+z_d} )
}
The kernel width $\sigma$ was chosen as the average of $\chi^2$ distances computed over a subset of the training set.
The normalization constant of the kernel is chosen as the standard deviation in Hilbert space, that is:

\eqf{
c &= \frac{1}{n}\sum_{i=1}^n K(x_i,x_i) - \frac{1}{n^2}\sum_{i=1}^n \sum_{j=1}^n K(x_i,x_j) \\
K & \mapsto \frac{K}{c} 
}
The constant was computed over the kernel from the training samples. An HIK kernel results in structures with finer resolution but less smoothness. Given the large morphological variability of cancer nuclei, higher smoothness is desired, and thus the $\chi^2$-kernel reflects better the available biological prior knowledge.

\subsection{Classifier: Learning algorithm for cancer evidence prediction}

A dual SVM from the shogun toolbox \cite{SonRaeHenWidBehZieBonBinGehFra10new} was used to predict the data from the averaged BoW feature. The only tunable parameter is the regularization constant of the SVM. The SVM regularization constant was not optimized and kept $C=1$ reflecting the normalization and prior knowledge from related datasets.

\subsection{Classifier: Experimental setup for cancer evidence prediction}

We divided the set of labeled patches into a training set of size $4142$ and an evaluation set of $4145$ patches. The negatively labeled patches contained a wide set of structures, including lymphocytes, myoepithelial cells, fibroblasts, and connective tissue without nuclei.

\subsection{Heatmapping: Cancer evidence prediction and heatmapping}

The heatmap for the cancer presence indicates localized evidence for presence of cancer-associated structures (ascending from green to yellow to red) versus evidence against presence of cancer-associated structures (ascending from green to blue). Neutral regions without any particular evidence are colored in green. A typical example of evidence against cancer-associated structures are lymphocytes or fibroblasts.

We followed the approach described for the $\chi^2$-kernel in \cite{bach2015pixel} for computing heatmaps. The most notable difference to the case of the HIK kernel is that we used the Taylor expansion of the $\chi^2$-kernel around a certain root point (cf. \cite{DBLP:journals/pr/MontavonLBSM17}). We will describe this in more detail now. The SVM prediction with multiple kernels $k_u$ for an image tile can be written as:
\be
f(x)=b+\sum_{i=1}^S\alpha_iy_i \sum_{u=1}^K \beta_u k_u(x_i,x)
\ee
We remind the reader about the used notation. The input $x$ for classification of an image tile is a concatenation of BoW features where $x_{d,u}$ is the $d$-th dimension of the $u$-th BoW feature in the concatenation. A BoW feature is computed as a sum of mappings $m$ of local features onto visual words. $m_d^{(u)}$ is the mapping onto the $d$-th dimension of the $u$-th BoW feature.

\be
x_{d,u}= \sum_{i=1}^N m_d^{(u)}(l_i)
\ee
Each local feature $l$ is used only for one BoW feature where $T(l)$ is the index of that BoW feature. Again as a first step, we need to compute a relevance score $r_{d,u}(x)$ for the dimension $d$ of the $u$-th BoW feature. Unlike with the HIK-Kernel, no direct decomposition exists for the $\chi^2$-kernel. The first step can be achieved by using first order Taylor expansions of the SVM prediction function  $f(x)$ for the BoW features $x$.

\begin{align}
\label{eq:taylorspec}
f(x) &\approx f(x_0) + \langle x-x_0, \nabla_x f(x_0) \rangle \nonumber \\ 
f(x) &\approx \sum_{u=1}^K  \sum_{d=1}^{V_u} \beta_u \left(\frac{f(x_0)}{V_u\|\vec\beta\|_1} + (x-x_0)^{(d,u)} \sum_{i=1}^S\alpha_iy_i \frac{\partial k_u(x_i, \cdot)}{\partial x^{(d,u)}}(x_0) \right) \nonumber \\  & = \sum_{u=1}^K  \sum_{d=1}^{V_u} r_{d,u}(x)
\\
r_{d,u}(x) & := \beta_u \left(\frac{f(x_0)}{V_u\|\vec\beta\|_1} + (x-x_0)^{(d,u)} \sum_{i=1}^S\alpha_iy_i \frac{\partial k_u(x_i, \cdot)}{\partial x^{(d,u)}}(x_0) \right),
\end{align}
where we select $x_0$ to be a point on the prediction boundary $f(x_0)=0$ and $V_u$ is the dimensionality of the $u$-th BoW feature. In case of kernels which are a sum of contributions along input dimensions such as the HIK, the Taylor expansion can be replaced by the sum across input dimensions as in \cite{DBLP:journals/ijcv/UijlingsSS12}. 

Our approach generalizes an idea from \cite{DBLP:journals/ijcv/UijlingsSS12} to non-linear kernels and general codebook mappings which we need for our application. 
\cite{caicedo1},\newline\cite{caicedo4} marked regions in histopathology images corresponding to a subset of most relevant visual words by a measure based on linear correlations between visual words and predictions. \cite{DBLP:conf/miccai/Cruz-RoaGGJEBMR12} computes class probabilities of visual words for BoW in a pLSA-like approach and projects them to local features using default nearest-neighbor mapping. 

From a purely methodical point of view, our contributions over these preceding works are two-fold. Firstly, our approach permits for the first time to use arbitrary local feature mappings. It is known from \cite{DBLP:conf/eccv/GemertGVS08,DBLP:conf/cvpr/WangYYLHG10} that more distributed local feature mappings like the rank-mapping employed here yield considerably better classification results than the nearest-neighbor mapping used in all preceding works cited above \cite{DBLP:journals/ijcv/UijlingsSS12,DBLP:conf/cvpr/LiuW12,caicedo1},\newline\cite{caicedo4,DBLP:conf/miccai/Cruz-RoaGGJEBMR12}. Secondly, we are enabled to use arbitrary differentiable kernel functions like the $\chi^2$-kernels applied here which yield superior performance over other kernels. 

Since we are interested in the explanation of the prediction of our system even in the case where the prediction contradicts human experience or manual annotations, we perform the Taylor expansion around a point $x_0$ which is a root of the prediction function $f(x_0)=0$. Note that for a linear kernel this expansion is exact. 

Since the set of root points is in the smooth case a manifold of codimension 1, the choice of $x_0$ is not unique. In order to obtain one root point $x_0$, we used the following approach: Take $C=30$ image tiles from a test set such that they have opposite sign of the SVM prediction for their set of BoW features $v_i, i=1, \ldots, C$ relative to $x$, i.e.~$f(x)f(v_i)<0$. Since the SVM prediction function is smooth, we know that the line $l(\alpha)= \alpha x + (1-\alpha)v_i$ must contain at least one root point $x_{i,0}$ of the SVM prediction function: $f(x_{i,0})$. Thus, line intersection methods yield $C=30$ root candidate points $x_{i,0}$. Now we choose a root candidate point. In this work the choice was based on minimizing the kernel-induced Hilbert space distance between the root candidate point $x_{i,0}$ and the prediction point $x$:
\be
x_0 = \mathrm{argmin}_i  \sum_{u=1}^K \beta_u \left( k_u(x,x) -2 k_u(x,x_{i,0}) + k_u(x_{i,0},x_{i,0}) \right),
\ee  

Let $r_{d,u}(x)$ be the relevance score for BoW feature dimension $d$ of BoW feature $u$ in an image tile. From here we continue as for the molecular profile prediction, except with the difference, that the relevance score is already a relevance score for one tile (because the prediction is made for each tile separately), so spreading it across image tiles is not necessary. We define a score for local features $R(l)$ via
\be
R(l):= \sum_{d \not\in Z_{T(l)}(x)  } r_{d,T(l)}(x) \frac{m_d^{(T(l))}(l)}{\sum_{i \mid T(i)=T(l) } m_d^{(T(l))}(i)} + \sum_{d \in Z_{T(l)}(x)  }  r_{d,T(l)}(x)\frac{ 1 }{\sum_{i  \mid T(i)=T(l) } 1}, 
\ee
where $Z_k$ is the set of dimensions onto which no local feature is mapped:
\be
Z_{k}(x)=\left\{d \mid \sum_{i \mid T(i)=k} m_d^{(k)}(i)=0 \right\}
\ee
We assume here that $0/0=0$. Finally the relevance of a pixel is the sum of the relevances scores over all local features which cover that pixel taken from one sub-image.
\be
rel(p)= \sum_{l \mid p \in \mathrm{support(l)}} R(l) \label{eq:unnormpixscores2}
\ee
For obtaining the heatmap of an image tile we normalize the scores from by the maximal absolute value of all scores:
\be
hm(p)=\frac{rel(p)}{\max_{s \in \mathrm{Tile}} |rel(s)| }\label{eq:normheatmap2}
\ee 
Eq. \ref{eq:normheatmap2} yields a score for every pixel in a tile.The final per-pixel scores are obtained by taking the mean from the relevance over all tiles containing this pixel. After that the scores are normalized to lie in the interval $[-1,+1]$.

\section{Supplemental Material: Lymphocyte evidence prediction and heatmapping}
\label{sec:lym}

Similar to the cancer evidence prediction task, the approach consists of two steps. The main differences are: different labels, a different feature set comprising multiple features, which is more suitable for lymphocytes, and the usage of the HIK kernel (all reflecting prior knowledge). In the first step we train a SVM classifier and use it to output a real-valued score indicating the confidence of a lymphocyte in one patch. In the second step the prediction score per patch is decomposed into a score per pixel using LRP. The scores of all pixels of one patch are then fused across patches and normalized to lie in the interval $[-1,+1]$. Mapping the scores onto colors yields then a heatmap for the full image providing visual representations lymphocytes (colors orange to red) and other cell types (shades of blue).

\subsection{Classifier: Labels for lymphocyte evidence prediction}

The label for each tile in the training and evaluation set was binary, $+1$ if it contained at least one lymphocyte cell, and $-1$ otherwise. We did not create any pixel-wise labels.

\subsection{Classifier: Parameters of the BoW Features used for lymphocyte evidence prediction}
At first an image was divided into tiles of size $102 \times 102$. The tiling was created with a stride of $102/3=34$ moving along a regular grid. We computed three BoW feature types per tile. SVM prediction and heatmapping were also done for each tile separately. Finally the heatmaps per tile were overlaid over the whole input image.

The local features are gradient norm quantiles, color intensity quantiles and SIFT. Their structure is described in detail in Section \ref{sec:bow}. Here we provide details on the parameters used for their computation. The local features were computed over red, green and blue channel, and then concatenated for these three channels. We sampled the local feature on a grid of step size $3$. We used scales $1.5$ and $2.0$  corresponding to local feature radii of $9$ and $12$ . This constitutes $3 \times 2$ BoW features. For the quantile based features we used a vocabulary size of $384$, for the SIFT feature $510$. In Section \ref{ssec:bowforcancer} we provide a rationale for not finetuning vocabulary sizes experimentally. 
From our experience with BoW features for the histopathology tasks, typically a vocabulary size of 384 to 600 is a good choice for for the SIFT local feature. Vocabulary size of quantile-based features was chosen as 384. This is due to the fact that the quantile-based local features had a low dimensionality, namely 18 per color channel, compared to 128 dimensions of SIFT. This follows our experience that comparably lower dimensional local features require also lower visual vocabulary sizes to cover the space of the local features. 

The final BoW feature was computed by mapping the local features by rank-weighted soft mapping, summing the resulting mapping vectors and normalizing them to unit $\ell_1$-norm.

\subsection{Classifier: Kernel for lymphocyte evidence prediction}
We used a histogram intersection kernel (HIK): 
\eqf{
K(x,z)= \sum_d \min(x_d,z_d)
}
The normalization constant of the kernel is chosen as the standard deviation in Hilbert space, that is:
\eqf{
c &= \frac{1}{n}\sum_{i=1}^n K(x_i,x_i) - \frac{1}{n^2}\sum_{i=1}^n \sum_{j=1}^n K(x_i,x_j) \\
K & \mapsto \frac{K}{c} 
}
The constant was computed over the kernel from the training samples. The HIK kernel gives finer-resolved heatmaps with structures closer to the size of a lymphocyte. We also tested a $\chi^2$-kernel which yielded slightly inferior performance.

\subsection{Classifier: Learning algorithm for lymphocyte evidence prediction}

A dual SVM from the shogun toolbox \cite{SonRaeHenWidBehZieBonBinGehFra10new} was used to predict the labels from the averaged BoW feature. The only tunable parameter is the regularization constant of the support vector machine. The SVM regularization constant was not optimized and fixed to $C=1$, because experience with other histopathological data sets showed little impact of the exact value of $C$.

\subsection{Classifier: Experimental setup for lymphocyte evidence prediction}
During test time we made for each tile at first a real-valued prediction of lymphocyte-presence by an SVM, and then computed pixel-wise scores for lymphocyte evidence for each tile separately. We used the same split of labeled patches into a training set of size $4142$ and an evaluation set of $4145$ patches a for cancer evidence. The negatively labeled patches contained a wide set of structures, including cancer cells, myoepithelial cells, fibroblasts, and connecting tissue without nuclei.

\subsection{Heatmapping: Lymphocyte evidence prediction and heatmapping}

The heatmap for the lymphocyte presence indicates localized evidence for presence of lymphocyte-associated structures (ascending from green to yellow to red) versus evidence against presence of lymphocyte-associated structures (ascending from green to blue). Neutral regions without any particular evidence are colored in green.
A typical example of evidence against lymphocyte-associated structures are cancer cells.

We followed the approach described for the HIK-kernel in \cite{bach2015pixel} for computing heatmaps. As before, the SVM prediction for an image tile and multiple kernels $k_u$ can be written as:
\be
f(x)=b+\sum_{i=1}^S\alpha_iy_i \sum_{u=1}^K \beta_u k_u(x_i,x)
\ee
We repeat the notation. The input $x$ for classification of an image tile is a concatenation of BoW features where $x_{d,u}$ is the $d$-th dimension of the $u$-th BoW feature in the concatenation. A BoW feature is computed as a sum of mappings $m$ of local features onto visual words. $m_d^{(u)}$ is the mapping onto the $d$-th dimension of the $u$-th BoW feature.
\be
x_{d,u}= \sum_{i=1}^N m_d^{(u)}(l_i)
\ee
The relevance score $r_{d,u}(x)$ for the dimension $d$ of the $u$-th BoW feature is computed as before:
\eqf{
k_u(x_i,x) &= \sum_{d=1}^{D_u} \min(x_i^{(d)},x^{(d)}) \\
\Rightarrow f(x) &=b+\sum_{i=1}^S\alpha_iy_i \sum_{u=1}^K \beta_u \sum_{d=1}^{D_u} \min(x_i^{(d)},x^{(d)}) \\
& = \sum_{u=1}^K \sum_{d=1}^{D_u} \frac{b \beta_u}{\sum_{u'} \beta_{u'} D_u}+ \beta_u\sum_{i=1}^S\alpha_iy_i \min(x_i^{(d)},x^{(d)})\\
f(x) & = \sum_{u=1}^K \sum_{d=1}^{D_u}r_{d,u}(x)\\
\Rightarrow r_{d,u}(x) & = \frac{b \beta_u}{\sum_{u'} \beta_{u'} D_u}+ \beta_u\sum_{i=1}^S\alpha_iy_i \min(x_i^{(d)},x^{(d)})
}
Similarly as for the cancer presence prediction task, this relevance score is already a score per tile. From here we compute a heatmap using equation \eqref{eq:locfeatformula2}, \eqref{eq:unnormpixscores}, \eqref{eq:normheatmap}. Fusing scores across tiles is done in the same manner as for the cancer prediction task.

\section{Supplemental Material: Histomorphological classification results}
\label{sec:spatclas}

\begin{figure}[h]
\centering
  \includegraphics[width=\textwidth]{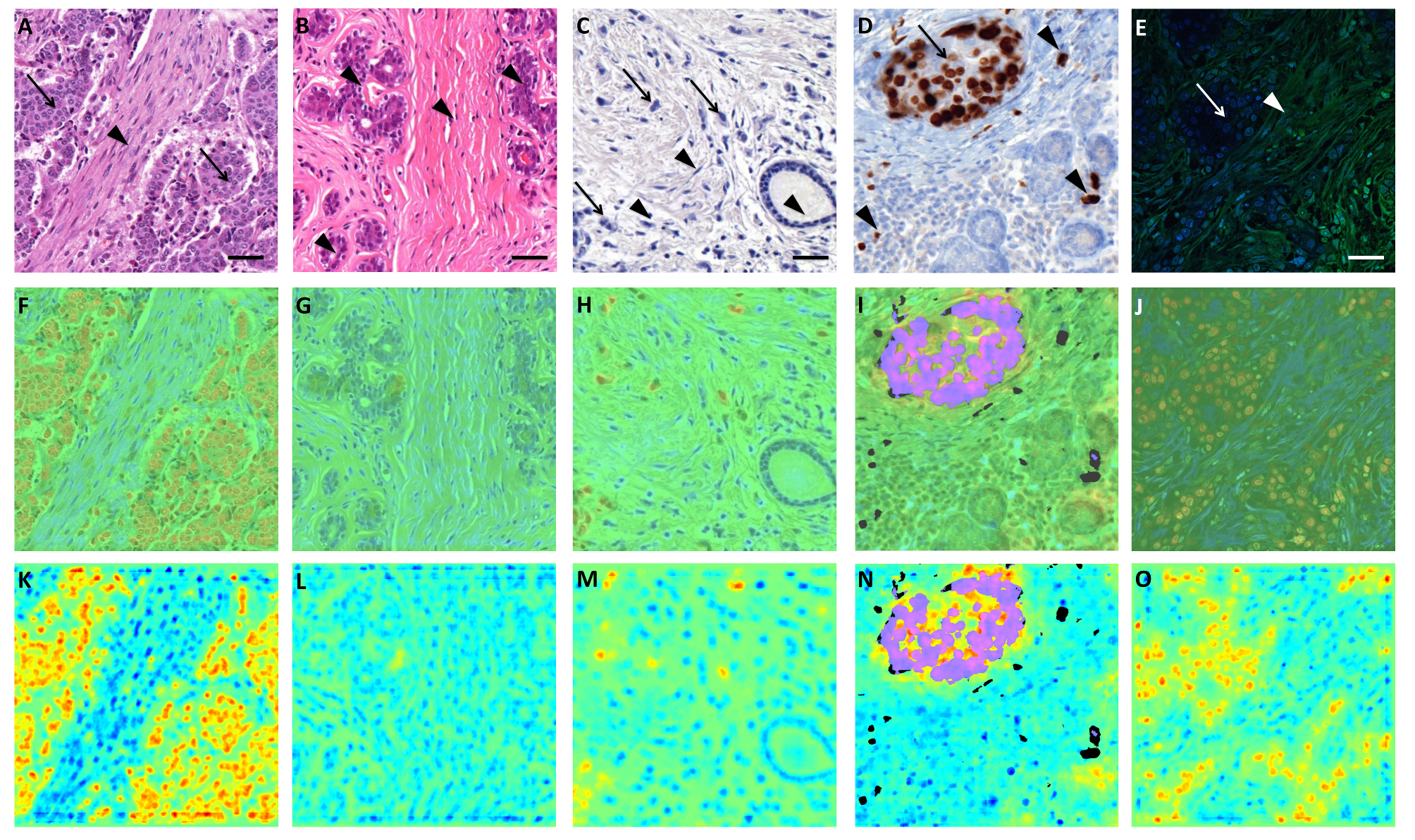}
\caption{\label{fig:suppl_spatfig}Spatial classification results. A-E: Original images; A: Invasive breast cancer, H\&E stain; B: Normal mammary glands and fibrous tissue, H\&E stain; C: Diffuse carcinoma infiltrate in fibrous tissue,
Hematoxylin stain; D: Diffuse carcinoma infiltrate next to normal mammary ducts plus immunohistochemistry for Ki67 proliferation scoring, Hematoxylin/DAB stain; E: Breast cancer and fibrous tissue, confocal fluorescent images
with nuclear and cytoplasmic dyes. \textit{Long arrow}: cancer cells; \textit{Short broad arrow}: non-cancer cells; \textit{Scale bar}: 50~$\mu$m. F-J: Corresponding classification result (absolute classification scores overlaid onto grayscale original image,
shades from  \textit{red} = cancer via \textit{green} = neutral to \textit{blue} = non-cancer indicate classification score; I, N: \textit{magenta}: Ki67-positive cancer cells, \textit{black}: Ki67-positive normal ducts and lymphocytes); K-O: Corresponding scaled heatmap.}
\end{figure}

We provide examples of histomorphological classification into tumor and healthy tissue on H\&E as well as on confocal images of breast cancer in Figure~\ref{fig:suppl_spatfig}. The first row displays original histopathological images of different staining procedures and different subtypes of cancer. For a good visual comparison heatmaps, obtained from LRP, are overlaid with the grey scaled original image. Furthermore the last row displays the plain heatmaps.

\noindent Deduced from the computed heatmaps, we computed pixel-wise accuracies for cancer and non-cancer cells. In particular, we took a radius of 10 pixels around the available point annotation of the cell center and derived pixel-wise accuracies in this way for the three different imaging modalities studied here, namely hematoxylin \& eosin (H\&E), nuclear staining (H) alone and fluorescence microscopy (FLUO) (cf. Table~\ref{tab:acc}). Furthermore, we investigated how these heatmaps scores vary with distance to cell centers, as marked by the point annotation. This is displayed, again for all three modalities, in Fig. 3, whereby yellow represents scores associated with cancer cells and blue those associated with non-cancer cells.
\begin{table} [h]
\centering
\caption{\label{tab:acc} Pixel-wise accuracies within a radius of 10 pixels of a cell center}
\begin{tabular}{|c|c|c|c|}
\hline
Imaging Modality &H\&E &  H&Fluorescence \\ \hline
Pixel-wise accuracies cancer cell&	84.72	&93.03&	78.65 \\ \hline
Pixel-wise accuracies non-cancer cell&84.03&	93.61	&71.55 \\ \hline

\end{tabular}
\end{table}

\begin{figure}[h]
\centering
  \includegraphics[width=\textwidth]{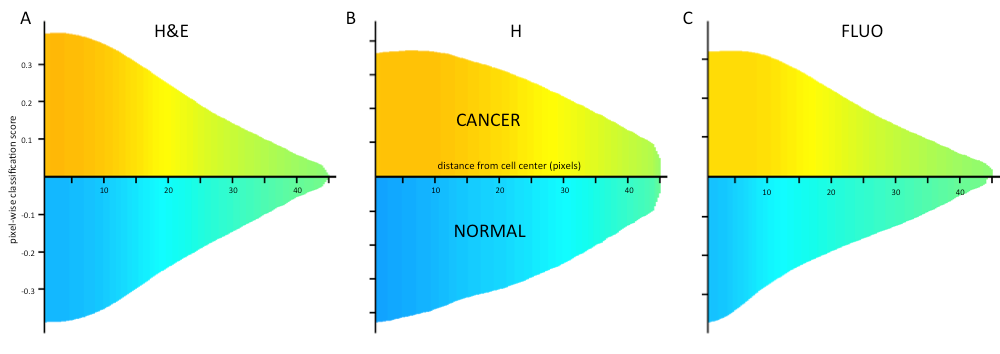}
\caption{\label{fig:dist}Visualization of pixel-wise classification accuracy as a function of the distance from cell centers in pixels.}
\end{figure}

\section{Supplemental Material: Survival time analysis}
\label{sec:survtimeanalysis_new2018}

We used the same BoW over SIFT feature and kernel as for molecular profile prediction in Section \ref{sec:mol}, however this time with survival time labels, which were available for $563$ patients. For training we used only the uncensored samples. We trained a binary SVM with HIK Kernel where a sample was labeled positive if survival time was above $60$ months, and negative otherwise.

\begin{figure}[h!]
\centering
\includegraphics[width=\textwidth]{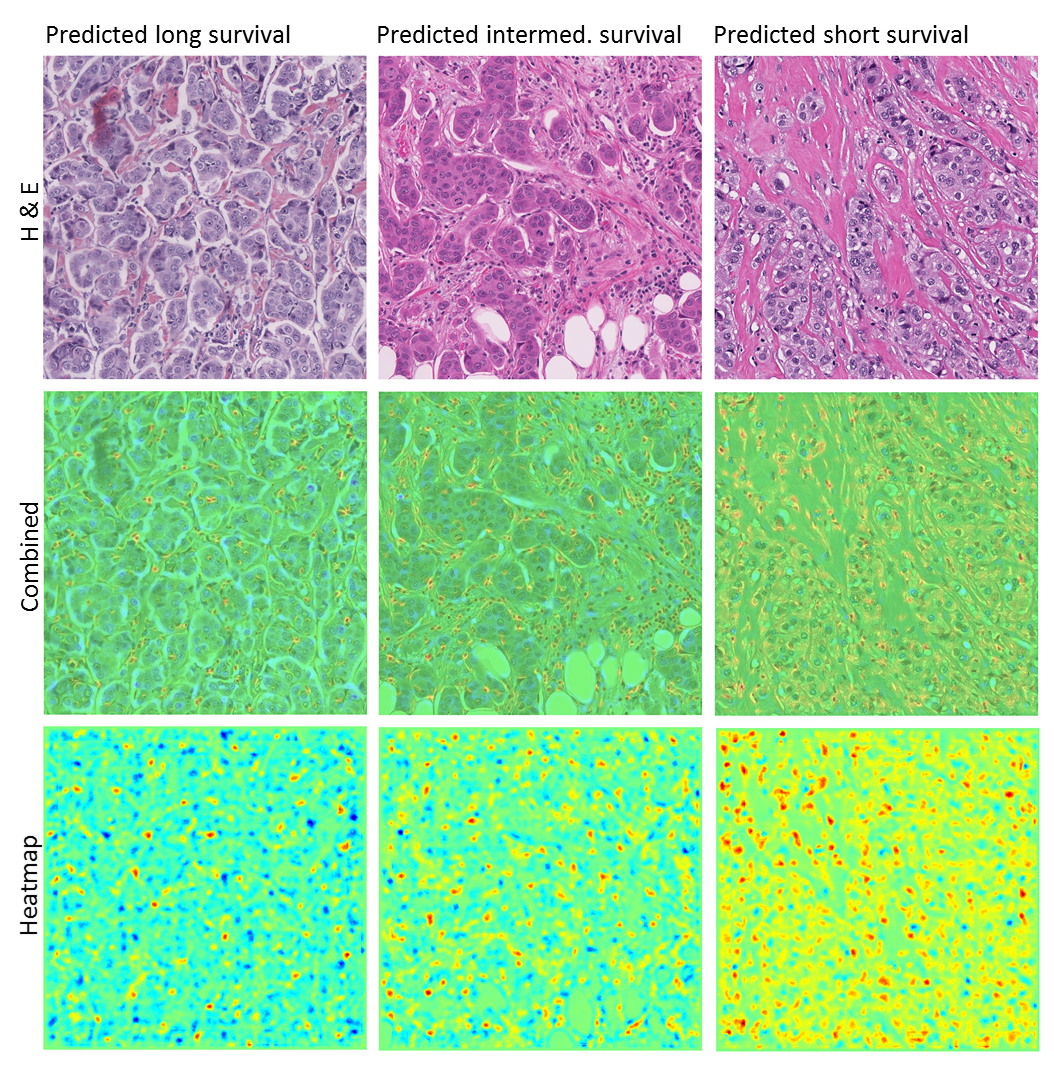}
\caption{\label{fig:survhm} H\&E images alongside with heatmaps representing pixel-wise contribution to survival prediction as well as a combined image overlaying heatmaps with H\&E images are shown for three hormone receptor positive breast carcinomas. Heatmap colors indicate regions associated with a prediction of prolonged survival (blue), shortened survival (red) as well as neutral regions (green). While it can be seen that cancer cells differ with respect to their contribution to prognosis between long, intermediate and short survival, tumor-infiltrating lymphocytes are predicted to contribute to shortened survival in all the shown luminal-type breast cancers, which is in line with initial clinical evidence we reported recently (Denkert et al., Lancet Oncol., 2018.)}
\end{figure}

\begin{figure}[hbt!]
\centering
\includegraphics[width=.66\textwidth]{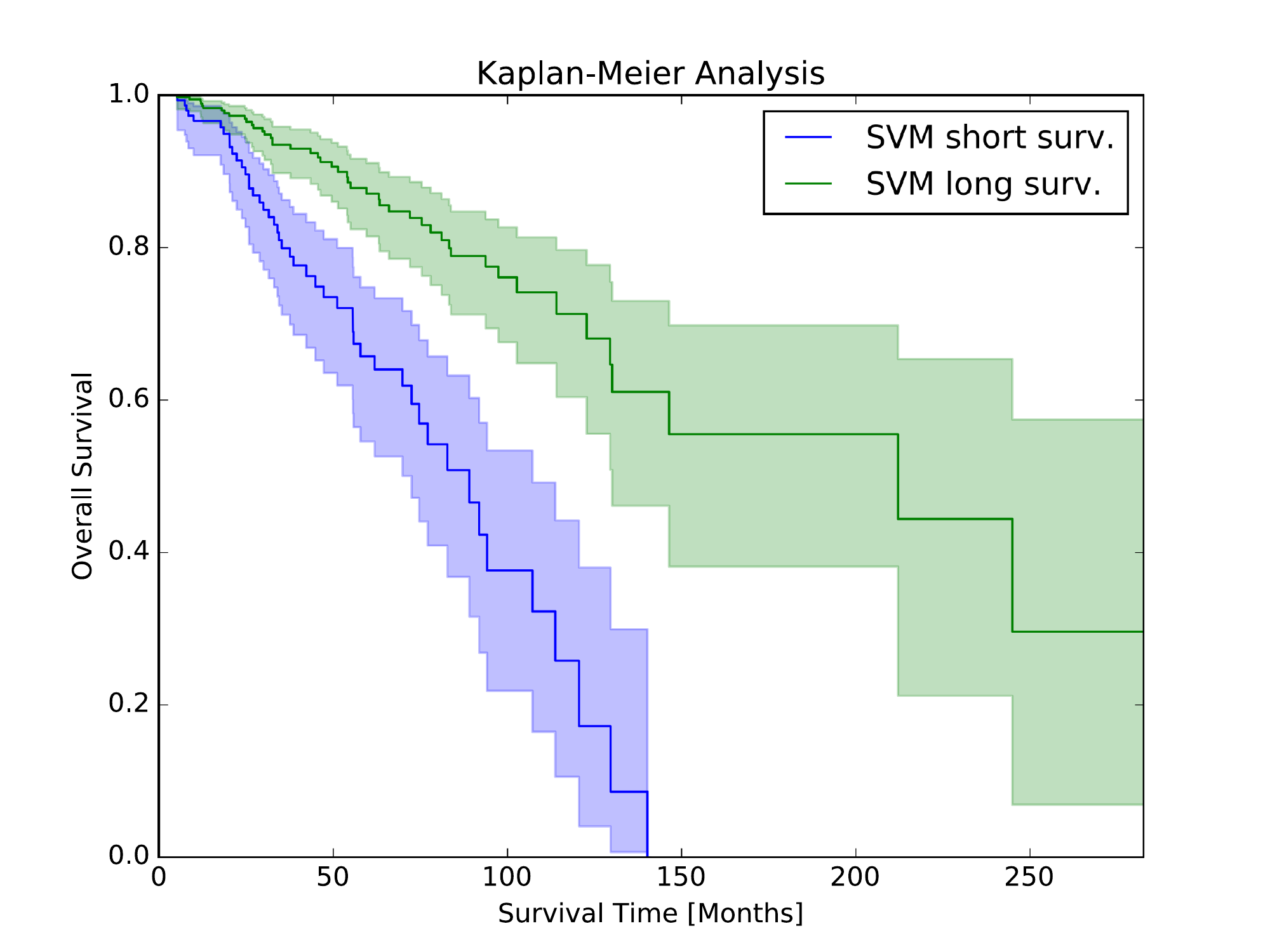}
\caption{\label{fig:kmplot} Kaplan-Meier Plot of the Survival Time Analysis. Log-Rank Test is significant with $p$-value $<6\cdot 10^{-10}$. Survival time of predicted low population is $41.4\pm 29.68$ months, survival time of high population is $47.12\pm 42.81$ months. Considering survival time averages restricted to the uncensored data only, on observes that the predicted low population of the uncensored ones has $49.45 \pm 34.39$, the predicted high population has $62.48 \pm 50.06$ months.  }
\end{figure}

As for corresponding images we only had 94 samples with uncensored labels. In order to make best use of this small sample size, we used again a 10-fold cross-validation, predicted for every cross-validation step on the withheld test split by a classifier which was not trained over the test split, and pooled the predictions over all test splits from the cross-validation. This problem does not appear for the censored cases, as they are excluded from the training. For the censored samples we have chosen for every sample one classifier randomly to predict on them, thus placing them randomly in one of the test splits.\\

\noindent For testing we performed the following test protocol:
\bi{
\item Uncensored samples: use prediction on test set for every cross-validation step. Average predictions over all test sets in the 10-fold cross-validation.
\item Censored samples: randomly select one of the classifiers for every censored sample. This corresponds to putting every censored sample randomly in one of the test splits. One could however also use the average of predictions of all 10 classifiers obtained from the cross-validation due to the fact that none of the censored samples was used in training. Although the latter mode yielded a better $p$-value, we report here the version with the random classifier to avoid a different handling of censored and uncensored samples at testing time.
}
 Heatmaps were generated using the same approach as for molecular feature prediction. In Figure~\ref{fig:survhm}, H\&E images alongside with heatmaps representing the contributions of the different images parts to survival as well as a combined image overlaying heatmaps with H\&E images are shown for three hormone receptor positive breast carcinomas. Heatmap colors indicate regions associated with a prediction of prolonged survival (blue), shortened survival (red) as well as neutral regions (green). Tumor infiltrating lymphocytes can be identified as negative prognostic factors, which is in line with findings we recently reported in a large clinical trial \cite{Denkert2018TumourinfiltratingLA}. Subsequently a Kaplan-Meier fit and a log-rank test (two-sided) were performed. Results can be seen in Figure \ref{fig:kmplot}.


\section{Supplemental Material: Predictions of hormone receptor status and tumor grading.}
\label{sec:hrprerher2grade}
In this section, we use our approach to prededict the most relevant clinicopathological features, i.e. the hormone receptor (HR) status including the progesterone (PR) and estrogen receptor (ER) status as well as the histopathological tumor grading from H\&E tumor slides. Again, we used the same BoW over SIFT feature and kernel as for molecular profile prediction in Section \ref{sec:mol}, but limited the regularization constant to $C=1$. 

In order to shed light on potential applications of these classifiers, we performed an analysis of the accuracy of predictions for tail sets under the SVM prediction $f(x)$, that is, we measured for a value $t>0$ the accuracy in the set of samples $S_-(t):=\{x:  f(x) \le -t\}$ and $S_+(t):=\{x:  f(x) \ge t\}$. We can show that several of the classifiers for HR, PR, ER and grading predictions have high accuracies of above $0.9$ for either positive or negative tail sets, even when the overall performance is about $0.63$ to $0.65$. Note that the assignment to tail sets is independent of labels and only depends on the SVM output. Therefore this criterion is applicable during deployment of a classifier.

The result regarding tail set accuracies can be explained by the nature of error distributions of classifiers: The errors in a trained classifier are often not spread out uniformly as a function of prediction scores. A trained classifier can be observed in practice to make less errors far away from the threshold of maximal uncertainty. This threshold would be $0$ for SVMs used here, and e.g.~$0.5$ for probabilistic two-class classifiers such as logistic regression. We observe the following results:
\ben{
\item \textbf{Grading}: We predicted two classes: Grade 1 and 2 versus Grade 3. We observed a balanced accuracy of $0.65 \pm 0.08$ and an AUC score of $0.72$. Tail scores: the accuracy is $\mathbf{0.91}$ for samples with score below $t=-1.0$ ($n=54$), the accuracy is $\mathbf{0.93}$ for samples with score above $t=1.04$  ($n=15$). The balanced accuracy is statistically significantly different from $0.5$ ($p=3.7\cdot 10^{-7}$).

\item \textbf{HR}: We observed a balanced accuracy of $0.64 \pm 0.09$ and an AUC score of $0.73$. The balanced accuracy is statistically significantly different from $0.5$ ($p=4.1\cdot 10^{-5}$). Tail scores: the accuracy is $\mathbf{0.9}$ for samples with score above $t=0.78$ ($n=130$) and $\mathbf{0.98}$ for samples with score above $t=1.3$ ($n=44$) .

\item \textbf{PR}: We observed balanced accuracy $0.59 \pm 0.07$ and an AUC score of $0.66$. The balanced accuracy is statistically significantly different from $0.5$ ($p=0.006$). Tail scores: the accuracy is $\mathbf{0.80}$ for samples with score above $t=0.82$ ($n=100$).

\item \textbf{ER}: We observed balanced accuracy $0.65 \pm 0.06$ and an AUC score of $0.7390$. The balanced accuracy is statistically significantly different from $0.5$ ($p=5\cdot 10^{-6}$). This result can be compared to the AUC score of $0.72$ in \cite{Rawat540} which uses a more complex pipeline consisting of segmentation and deep learning over segmented nuclei. Tail scores: the accuracy is $\mathbf{0.9}$ for samples with score above $t=0.84$ ($n=110$) and $\mathbf{0.96}$ for samples with score above $t=1.08$ ($n=67$) .

\item \textbf{HER2}: We also attempted to predict clinical HER2-Status, which was, however, not significant ($c=0.55$, $p=0.46$).
}


\section{Supplemental Material: Validation of morphomolecular heatmaps}
\label{sec:synthmiriam}

\subsection{Validation of computational image prediction by comparison with immunohistochemistry} 

We use the Quadrat method from (geo-)spatial statistics \cite{gatrell1996} to test the hypothesis that in the machine learning-based morpho-molecular predictions, the molecular predictions are statistically significantly spatially associated with morphological structures and have non-random spatial distributions. With this method, the image region is subdivided into subtiles, for each of which the contained signal for morphological and molecular features are computed resulting in a 2x2 matrix with values indicating that both features are coincident, only one feature is present, or both are absent (cf. Table~\ref{tab:cont_idea}). $\chi^2$-statistics is used to test for significance of co-clustering of the morphological and molecular features. We used a tile size of 10\% of the image width.

\begin{table}[ht]
	\centering
	\begin{tabular}{c|c}
	number of image tiles that contain & number of image tiles that contain \\
	both morphological and molecular information & only morphological information\\ \hline
	number of image tiles that contain & number of image tiles that contain \\
	only molecular information & neither morphological nor molecular information
	\end{tabular}
	\caption{Contingency table example for the Quadrat test.}
	\label{tab:cont_idea}
\end{table}

\begin{figure}[htb]
  \centering
    \includegraphics[width=.72\textwidth]{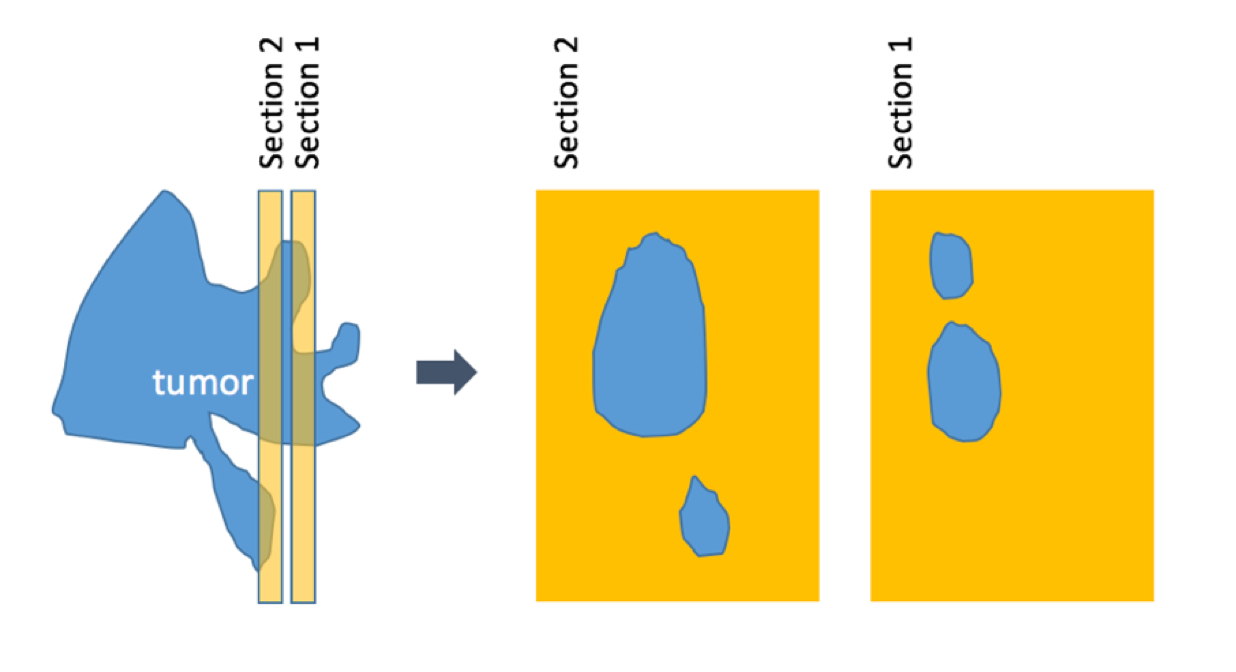}
    \caption{\label{fig:tumorstructure} Tumors often show complex 3D architecture which may lead to substantial morphological variation in consecutive 2D sections, making it difficult to directly compare one section (used for H\&E stain) and an adjacent section (used for IHC).}
\end{figure}

Although H\&E and IHC stain show visually similar structures, due to the often complex three-dimensional architecture of tumors and the need to use consecutive tissue sections to perform both H\&E and IHC, the images often have sufficiently pronounced differences that render direct quantitative and statistical comparisons subject to systematic tissue variability,  See Figure \ref{fig:tumorstructure} which provides a pictorial visualization for these differences. As a solution, to still provide a quantitative evaluation of the concordance between prediction and experiment, we first evaluate the statistical significance/non-randomness of the association of the predicted molecular features and morphological properties by using the Quadrat test. 

This analysis shows that the spatial associations observed between molecular markers and morphology in the computationally generated morpho-molecular images are highly statistically significant as they are in the IHC images. Second, we quantitatively compare for the predicted and experimentally measured images the numbers of image regions in which molecular and morphological properties do and do not spatially co-localize. This analysis shows that the ratios of spatially co-localized and non-co-localized image regions are similar comparing prediction and experiment for each marker, but are substantially different across markers (main manuscript Fig. 6). 

\subsection{Immunohistochemistry}
Immunohistochemical  stains  were  performed  on  formalin fixed,  paraffin embedded  tumor material according to standard protocols on an automated Ventana Benchmark HX IHC system (Ventana-Roche Medical  systems,  Tucson,  AZ,  USA)  following  the  manufacturer’s  instructions. Antibodies  against  p53  (DO-7, Dako), MSH2 (G219-1129, Ventana) and  E-Cadherin (4A2C7, Zytomed) were  used  in  a  dilution  of 1:50. Tumor material was obtained from the tissue archive of the Institute of Pathology of the Charit\'e Universitaetsmedizin Berlin, after patients had given their informed consent.

\subsection{Synthetic data experiment on the reliability of heatmaps}
\subsubsection{Aim of the synthetic data experiment}

The aim of this synthetic data experiment is to analyze the reliability of heatmaps also at relatively lower prediction accuracies. Secondly, we aim to show that increasing the amount of data leads to a higher balanced accuracy score and better quality of heatmaps but that nevertheless the main trends in heatmaps are already detectable and distinct at a low balanced accuracy score.\\ 
Synthetic data experiments allow to break down complex analyses on, although often well-controlled, likewise complex data. In particular, we were interested in replicating the prediction of molecular features from microanatomic images on a highly simplified and thus easier to interpret dataset. Therefore, we reapplied the classification pipeline of the main paper on an artificially generated dataset, from which we sampled three differently sized datasets, and evaluated contributions to the particular classifier's decision. Here the synthetic data experiment provides the unique chance to precisely formulate the expectations of a correctly made decision by virtue of knowing the underlying mechanism for generating the data. This expectation can be expressed and visualized through a ground truth heatmap and can be used as reference when evaluating the results.

\subsubsection{Synthetic data}
\paragraph{Images}

\begin{figure}[hbt]
\centering
\includegraphics[width=.6\textwidth]{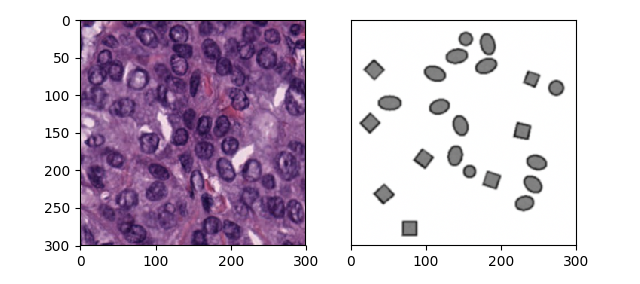}
\caption{\label{fig:toy_orig} Left: Patch of a histopathological image. Right: Simplified synthetic data image, where cell types are represented by various geometrical shapes. An artificial molecular score, which then in turn was used to define a binary molecular label per image, was created on basis of the contained shapes in the image.}
\end{figure}

The essential data to support our hypothesis need to resemble the highly complex histopathological images in the main relevant properties while at the same time simplifying those features that are not contributing primarily to the analysis. Precisely, on the basis of the assumption that morphologically different cell types can be discriminated to a large extend by their particular shape, the generated patches display geometrical shapes on a white background (cf. Figure \ref{fig:toy_orig}). For the experiment, we generated a total of 1500 square patches with an edge length of 300 pixels. Geometrical shapes include circles, ellipses and squares, and are depicted in gray with black edges. Mathematical properties, such as the size of shapes and the distribution of molecular scores were matched to data from the original dataset. The lengths and radii of shapes respectively were chosen from Gaussian distributions 
$\mathcal{N}(\mu_{s},\,\sigma_{s})$ with the following means $\mu_{s}$ and standard deviations $\sigma_{s}$ for the particular shapes:
\begin{eqnarray*}
&\mu_{s} = 20,\,\sigma_{s} = 1& \text{\quad for lengths of squares} \\
&\mu_{s} = 10,\,\sigma_{s} = 1& \text{\quad for radii of circles} \\
&\mu_{s} = [10,15],\,\sigma_{s} = [0,1]& \text{\quad for radii of ellipses}
\end{eqnarray*}
The total number of shapes in patches was drawn from a Gaussian distribution $\mathcal{N}(\mu_{n},\,\sigma_{n})$ with $\mu_{n}=15$ and $\sigma_{n}=2$. The underlying distribution of count of shapes, i.e. simulated morphology was based on circles as target shape. One third of the images contained no circles while in the other two thirds the ratio of the number of circles to the total amount of shapes was randomly chosen to be between 10\% and 100\%. In order to make the synthetic data resemble medical images, the blurring of medical images was imitated by a Gaussian filter with kernel size of 5x5. Furthermore, we altered every pixel by an additive, uniformly distributed noise factor $\epsilon \in [-1,1]$ to account for random noise information of camera sensors. 

\paragraph{Molecular scores}

As we were trying to replicate the results of the molecular analysis on a simpler dataset where the ground truth is known, we need to artificially generate a real-valued molecular score per image. This was achieved by calculating a weighted sum of the number of particular shapes present in a patch, i.e.
\begin{equation}
\label{eq:mol_score}
\text{molecular score} = \sum_{i=1}^{3} (a_{i} \cdot |s_{i}|+\epsilon_{i}) +\epsilon \qquad \epsilon_{i},\epsilon \in [0, 0.1)
\end{equation}
where $|s_{i}|$ denotes the number of a particular geometrical form in an image and $\epsilon$ represents the measurement noise. The weights $a_i$ need to be chosen cautiously in order to create matching ground truth heatmaps for the generated images and therefore allowing a precise evaluation of the results. Note that if all geometrical shapes were partially contributing to the molecular score, it would be difficult to visibly discriminate the relevances of particular shapes and therefore to assess the quality of heatmaps. Ideally, one would construct a molecular score that results in a ground truth heatmap in which differences in importance of the shapes’ contribution are clearly visible. Hence, we chose a molecular score where ellipses contributed twice (0.6) as much as circles (0.3) and squares did not affect the molecular score at all. In terms of heatmaps, we therefore predict positive evidence on circles and ellipses and neutral or negative evidence on the rest of the patch. To turn this into a classification problem, the real valued score was separated by the 50\% quantile of each dataset.

\subsubsection{Analysis}

In order to put the classification accuracy and the quality of heatmaps in relation to the number of training data, we performed all of the following analyses for training datasets that contained 100, 500 and 1500 images. Hereby the larger datasets emerged by adding new data to the existing smaller datasets. The following analysis is performed very closely to the analysis described in paragraph 2 of the supplemental material.\\
As before, for every image in the particular dataset SIFT features were extracted with a keypoint scale of 2.0 on a regular grid which was superimposed on the image whereby grid points were half a feature radius apart. Due to computation time, we only considered 30\% of features in about 30\% of images to compute the codebook of visual words, which for sufficient large amount of data will capture the most important features. In line with the original analysis, the codebook contained 510 visual vocabularies. Subsequently SIFT features were mapped onto the visual codebook by rank-weighted soft mapping 
and then averaged over locations. The resulting global feature per image was then L1-normalized.\\
Classification was performed by a support vector machine (SVM), whereby the regularization parameter C was determined from a 5-fold cross-validation with $C=4^{n}$, $n=\{-2,-1, ..,2\}$. In order to account for non-linear boundaries we applied a histogram intersection kernel to the data prior to classification. Evaluating the performance of the SVM was done by using the average of the true positive and true negative rate. The contribution of single pixels to the classifier decision was computed by back-propagating the outcome of the SVM by the layer-wise-backpropagation algorithm as described in \cite{bach2015pixel}.
An independently generated sample of 100 test images was used to compare between the different conditions, i.e. different amount of training data. For this subset of images we additionally generated "ground truth" images, in which the areas of circles and ellipses were denoted with 0.5 and 1 respectively and all other pixels were set to 0. This is based on the expectation that ellipses contribute twice as much as circles to the classifier decision. Additionally to computing balanced accuracy scores for the noisy training labels, we also evaluated performance on ground truth labels, which means that we neglected the measurement noise in eq (\ref{eq:mol_score}). This uses the unique chance of evaluating the performance of the trained SVM on the labels that we are generally interested in but that usually are unavailable to us due to observational noise. 

\subsubsection{Results}

\begin{table}[ht]
\centering
\begin{tabular}{cccc}
\hline
 & n=100 & n=500 & n=1500 \\ \hline
Noisy Labels & 0.75 & 0.79 & 0.82  \\
Ground truth labels & 0.79 & 0.9 & 0.9  \\
\hline
\end{tabular}
\caption{\label{tab:pred_score} Balanced accuracy scores on the validation dataset of the three models (using differently sized training datasets). The balanced accuracy scores were both evaluated for labels containing observational noise and for ground truth labels emerging directly from the generating mechanism.}
\end{table}

\begin{figure}[h]
  \centering
    \includegraphics[width=.9\textwidth]{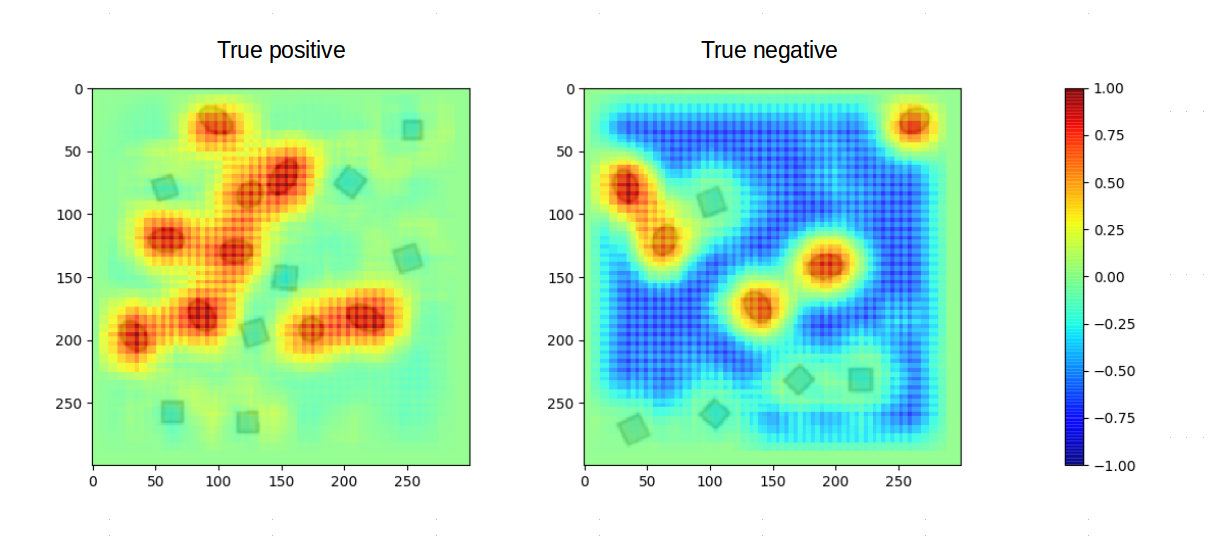}
    \caption{\label{fig:tp_tn} Exemplary heatmap of a true positive and a true negative, i.e. both images were correctly classified for the positive and negative class respectively. The heatmaps result from the classification decision of the the model with 100 training images. Note that in both cases the target shapes are prominent.}
\end{figure}

\begin{figure}[p]
  \centering
    \includegraphics[width=.9\textwidth]{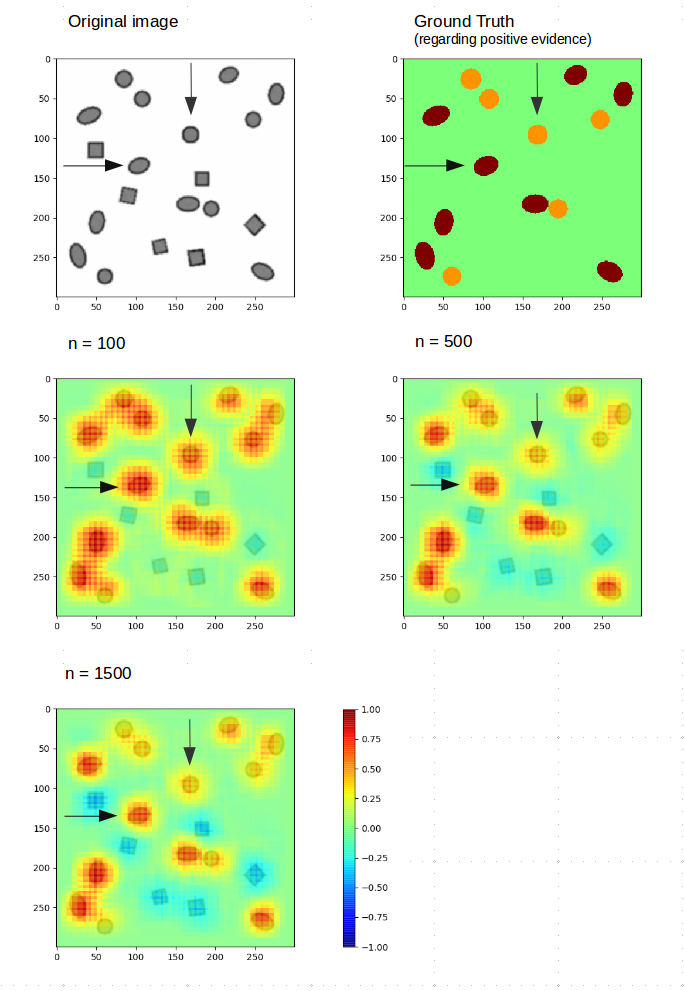}
    \caption{\label{fig:comparison} For reason of comparison, 
a representative example of the evolution of heatmaps over differently sized datasets (middle \& bottom row) is displayed as well as its ground truth (top right). The ground truth image illustrates the distribution of positive evidence as 	defined by the generating mechanism for the synthetic images. Arrows point to one example of the respective typical target image, i.e. ellipse and circle. The image was classified correctly by all three models.}
\end{figure}

\noindent In this synthetic data experiment, we tested three differently sized datasets to analyze the heatmap quality as a function of increasing number of images. Results of the classification for the different models are listed in Table\ref{tab:pred_score} and demonstrate that with increasing availability of training images the predictions of the models become more accurate. The comparison to the ground truth label of the validation images shows that the prediction error on noisy labels may seem very small (around 75\%-80\%) but that the model still predicts the underlying truth quite well (up to 90\%). Note however, that this can only point out a tendency in evolution of the heatmaps but one cannot conclude the number of necessary images from this experiment for medical data as these synthetic images are highly simplified.\\
A main goal of this synthetic data experiment was to show that also when the balanced accuracy score for molecular classes is rather low, the heatmaps still reveal relevant information about pixel contributions. Figure \ref{fig:tp_tn} displays a heatmap for each of the two separated classes, i.e. high versus low artificial molecular score, for the model trained on 100 images which has a balanced accuracy score of 75\%. It becomes apparent from the figure that the key elements are captured already with this rather low balanced accuracy score, i.e. positive relevance is predominantly distributed to the target shapes. Furthermore, as expected all target shapes contributed to the classifier decision. Note that in contrast to morphological classifications, where the presence of certain shapes leads to positive labels, here these target shapes always contribute to the positive evidence even if the image has a negative label. Negative predictions are achieved by outweighing positive by negative evidence (cf. right plot in Figure \ref{fig:tp_tn}). Negative evidence is mainly distributed over the background of the image due to the absence of relevant shapes in this part that could contribute to exceeding the labeling threshold.\\
If we compare heatmaps generated by models with different availability of data, one can observe that even though main elements are already captured in the model with 100 training images, the quality of the heatmap increases with more available training images in terms of focusing positive relevances on relevant shapes and the disparity of importance of circles and ellipses to the classifier decision (cf. Figure \ref{fig:comparison}). Moreover one can notice that  the relevance of the squares decreases with enlarging amount of training data, as the classifier learns that if a large ratio of the shapes are squares the probability for the positive class drops accordingly.
However, it seems that there is only a very subtle difference between the models with more data, which is in line with the equal prediction rate of these two models on the ground truth labels.\\
In conclusion, we could show that, also at a rather low balanced accuracy score, heatmaps contain relevant information for interpreting classifier decisions. Furthermore increasing availability of data will, until a certain point, not only improve the classification rate but also the quality of heatmaps.   

\paragraph{Statistical Testing}

To test for the alignment of morphology and molecular features, we make a Quadrat test on the morphological shapes, taken from the ground truth data, together with the computed molecular heatmaps. The Quadrat test, consisting of non-overlapping 50x50 pixels windows, was performed on the heatmaps created with n=100 training images. The statistic is performed over all test images (n=100). Thus we obtain a value of $\chi^2 = 2134.9$ with $p = 0.0$ (cf. Table \ref{tab:synth_contingency}), which confirms the results obtained by visual inspection of the heatmaps.

\begin{table}
\centering
	\begin{tabular}{lcc}
	\hline
	& molecular feature & no molecular feature \\ \hline
	tumor & 1428 & 8\\
	no tumor & 452 & 1712\\
	\hline
	\end{tabular}
	\caption{Contingency table, $\chi^2 = 2134.9$ with $p = 0.0$.}
	\label{tab:synth_contingency}
\end{table}


\section{Supplemental Material: Results of significance assessment of molecular feature prediction}
\label{sec:resultsstats}

In order to test for statistical significance of the reported balanced accuracy against random guessing and to estimate the false discovery rate (FDR) and correct for multiple testing, we performed the following two tests:
\begin{itemize}
	\item Statistical Testing and FDR control based on Hoeffding's inequality and Benjamini-Hochberg
	\item Monte-Carlo Estimation of the False Discovery Rate
\end{itemize}
The results of these tests for the different entities are reported below. Details of the tests, and where necessary mathematical derivations, are described extensively in Section \ref{sec:derivationstats}.

\subsection{Results of testing based on Hoeffding and Benjamini-Hochberg}

This approach is based on the idea to define a statistical test per gene or protein, which yields a $p$-value, stating if the computed balanced accuracy is significantly different from 0.5. Subsequently we correct for multiple testing and state the number of genes/proteins which are significant after FDR control. In general the approach is based on the statement of the Hoeffding inequality and Benjamini-Hochberg for FDR correction.

\subsection*{DNA methylation (METH)}
The balanced accuracies reported in the main paper are for $K=19953$ genes based on $n=400$ patients. For every patient we processed one image. $K$ is the relevant number for the FDR control, $n$ influences the distribution of the $p$-values for single genes/proteins. For a FDR-level $\alpha=0.05$ there are $9087$ statistically significant gene methylations. If we take a closer look at the intersection with sets of balanced accuracy above an exemplary threshold, then we find:
\bi{
\item for balanced accuracy $0.6 \le b < 0.62$: 2241 out of 3858 genes are significant. 
\item for balanced accuracy $\ge 0.62$: all 3705 genes are significant. 
}
A detailed list of results over all molecular features is available from the authors. The column {\tt significance} in it denotes significance when using Benyamini-Hochberg FDR with FDR threshold $\alpha=0.05$ and the assumption of no or positive correlations.

\subsection*{Copy number variation (CNV)}
For this $K=24475$ CNVs were tested on $n=555$ cases. As expected, due to the slightly larger sample size compared to DNA methylation, the fraction of significant results for smaller values of $\alpha$ is much larger and otherwise they are comparable. For a FDR-level $\alpha=0.05$ there are $12267$ statistically significant CNVs. If we take a closer look at the intersection with sets of balanced accuracy above an exemplary threshold, then we find:
\bi{
\item for balanced accuracy $0.55 \le b < 0.6$: 6993 out of 15763 genes are significant. 
\item for balanced accuracy $\ge 0.6$: all 5274 genes are significant. 
}
A detailed list of results over all molecular features is available from the authors. The column {\tt significance} in it denotes significance when using Benyamini-Hochberg FDR with FDR threshold $\alpha=0.05$ and the assumption of no or positive correlations.

\subsection*{Protein expression (PROT)}

Here $n=565$ and $K=190$.
\bi{
\item for balanced accuracy $ 0.57 \le b < 0.62$: 14 out of 45 proteins are significant (and one more with balanced accuracy of $0.569$).
\item for balanced accuracy $\ge 0.62$: all 4 proteins are significant.  
}
Overall PROT seems to be a notably harder problem to predict than all the others. This can be also seen by the fact that only 4 proteins out of 190 are above $0.62$ in balanced accuracy and might be explained by higher amounts of noise in the protein data. A detailed list of results over all molecular features is available from the authors.

\subsection*{Gene expression (RNASEQ)}
For RNASEQ we performed $K=20530$ different tests on $n=563$ images. They are stronger than for CNV because the number of images $n$ is comparable, but less tests were performed.For a FDR-level $\alpha=0.05$ there are $12635$ statistically significant RNAs. If we take a closer look at the intersection with sets of balanced accuracy above an exemplary threshold, then we find:
\bi{
\item for balanced accuracy $0.55 \le b < 0.6$: 5559 out of 10074 genes are significant. 
\item for balanced accuracy $\ge 0.6$: all 7076 genes are significant. 
}
A detailed list of results over all molecular features is available from the authors. The column {\tt significance} in it denotes significance when using Benyamini-Hochberg FDR with FDR threshold $\alpha=0.05$ and the assumption of no or positive correlations.

\subsection*{Mutation data (SOM)}
We trained the machine learning approach for 23 genes which are mutated in at least 10 cases, but applied statistical testing only to the subset of those 10 genes that had at least 15 cases with mutations. Below the statistically significant findings are listed for all tested mutations. Interestingly, the 5 genes with lowest sample size are those with the lowest balanced accuracy, from MAP2K4 down to PTEN as seen below. The top 5 had all more than 30 positive samples. From those top five, four are statistically significant for $\alpha=0.05$. For $\alpha=0.005$ the first three are still statistically significant. 

\begin{table}[h!]
	\centering
	\begin{tabular}{l|llllllllll}
	Gene & \textbf{TP53} & \textbf{PIK3CA} & \textbf{CDH1} & \textbf{GATA3} & MLL3 & MAP2K4 & DST& NF1 & SYNE1 & PTEN \\
 	balanced acc. & 0.618 & 0.593 & 0.655 &  0.600 & 0.568 & 0.548 & 0.524 & 0.496 & 0.487 & 0.498
 	\end{tabular}
	\caption{Balanced Accuracies. Bold-faced are those with statistical significant outcome for $\alpha=0.05$}
\end{table}
\noindent The fact that PIK3CA has a lower balanced accuracy but a higher $p$-value than CDH1 is not a mistake. It comes from the fact, that a quantile closer to the median was used, which results in a higher effective sample size and thus lower $p$-value.
\begin{table}[h!]
	\centering
	\begin{tabular}{l|llll}
	Gene & TP53 & PIK3CA & CDH1 & GATA3\\
 	balanced accuracy & 0.618 & 0.593 & 0.655 &  0.600  \\
 	$p$-value with FDR & 9.06e-07 & 0.0002885 & 0.0003528 & 0.013573 \\
 	threshold from BH-procedure ($\alpha=0.05$) & 0.005 & 0.0100 & 0.015 & 0.02000
 	\end{tabular}
	\caption{Test results of Benjamini-Hochberg for FDR-level $\alpha=0.05$.}
\end{table}

\subsection{Results of Monte-Carlo estimation}
The application of Monte-Carlo Estimate to this problem setting yields an upper bound on the false discovery rate (FDR) referable to multiple testing. In short, we repeatedly create artificial gene/protein labels by allocating random labels and then compute the number of false discoveries for a given balanced accuracy. Below we report the FDR for DNA methylation (METH) for the computed balanced accuracies.

\subsection*{DNA methylation (METH)}
The Monte-Carlo estimate for METH was based on $m=1350$ independent runs with random permutations. Each run corresponds to one simulated gene and used the same $n=400$ patients as for the experiments with the real labels. For every patient we used one image, so $n=400$ images. We obtained as upper bounds on the FDR:
\begin{table}[h]
	\centering
	\begin{tabular}{l|lll}
	Balanced accuracy & $\ge$60 & $\ge$62 & $\ge$64 \\
	False Discovery Rate & 0.1547 &  0.0997 & 0.0467\\ 
	\end{tabular}
	\caption{Upper bound on FDR given the balanced accuracy}
\end{table}
Thus, for balanced accuracies of $\ge 60\%$ we obtain an FDR of below $15.5\%$, for balanced accuracies of $\ge 62\%$ we obtain an FDR of below $10.0\%$, and for balanced accuracies of $\ge 64\%$ an FDR of below $4.7\%$.

\subsection*{Copy number variations (CNV)}
For the estimation for CNV we used $m = 1000$ independent runs, and used the same $n=555$ patients as for the experiments with the real labels.
\begin{table}[h]
	\centering
	\begin{tabular}{l|lll}
	Balanced accuracy & $\ge$60 & $\ge$62 & $\ge$64 \\
	False Discovery Rate & 0.1174 &  0.0839 & 0.0216\\
	\end{tabular}
	\caption{Upper bound on FDR given the balanced accuracy}
\end{table}
\noindent Thus, for balanced accuracies of $\ge 60\%$ we obtain an FDR of below $11.7\%$, for balanced accuracies of $\ge 62\%$ we obtain an FDR of below $8.4\%$, and for balanced accuracies of $\ge 64\%$ an FDR of below $2.2\%$

\subsection*{Protein expression (PROT)}
For the estimation for PROT we used $m = 1000$ independent runs, and used the same $n=565$ patients as for the experiments with the real labels.
\begin{table}[h!]
	\centering
	\begin{tabular}{l|lll}
	Balanced accuracy & $\ge$60 & $\ge$62 & $\ge$64 \\
	False Discovery Rate & 0.4238 &  0.285 & 0.38\\
	\end{tabular}
	\caption{Upper bound on FDR given the balanced accuracy}
\end{table}

\subsection*{Gene expression (RNASEQ)}
The results for RNASEQ are based on $m = 1000$ independent runs and used the same $n=563$ patients as for the experiments with the real labels.
\begin{table}[h!]
	\centering
	\begin{tabular}{l|ll}
	Balanced accuracy & $\ge$60 & $\ge$62 \\
	False Discovery Rate &  0.0463 & 0.0157\\
	\end{tabular}
	\caption{Upper bound on FDR given the balanced accuracy}
\end{table}
\noindent Thus, for balanced accuracies of $\ge 60\%$ we obtain an FDR of below $4.6\%$, for balanced accuracies of $\ge 62\%$ we obtain an FDR of below $1.6\%$.


\section{Supplemental Material: Methods for significance assessment of molecular feature prediction}
\label{sec:derivationstats}

\subsection{Statistical testing and FDR control based on Hoeffding and Benjamini-Hochberg}

The reported results in the manuscript are empirical balanced accuracies $\overline{Bac}$: \footnote{Notation: 1[x=1] is 1 if the expression is true and 0 if it is false.}
\eq{
\overline{Bac}&=  \frac{1}{2}\frac{1}{n_+} \sum_{i=1}^n 1[y_i=+1] 1[f(x_i)=y_i] + \frac{1}{2}\frac{1}{n_-} \sum_{i=1}^n 1[y_i=-1] 1[f(x_i)=y_i]\\
&= \frac{1}{n}   \sum_{i=1}^n  1[f(x_i)=y_i] \left( \frac{n}{2n_+} 1[y_i=+1] + \frac{n}{2n_-} 1[y_i=-1] \right) \label{eq:balacc}\\
n_+&= \sum_{i=1}^n 1[y_i=+1] , \   n_-= \sum_{i=1}^n 1[y_i=-1] 
}
This is not a standard procedure and therefore needs some mathematical derivations. This is due to the fact that the reported balanced accuracy is a sample average and not a binary classification itself (for more detail see below). As seen in eq. \eqref{eq:balacc} the empirical balanced accuracy is a sample average of binary decisions but not a binary decision itself and therefore requires an adopted statistical test.

For each gene one can consider the true expectation $\mathbb{E}[Bac]$ of the empirical balanced accuracies $\overline{Bac}$. It is known that the random guessing threshold for the expectation $\mathbb{E}[Bac]$ is $0.5$ for binary classification. The statistical test now asks, given an observed empirical average $\overline{Bac}$, how probable is it, that the true expectation $\mathbb{E}[Bac]$ is $0.5$ or below. Therefore we formulate the null hypothesis as the scenario where the expected balanced accuracy $\mathbb{E}[Bac]$ is equal or smaller than the random guessing threshold, that is that we did not learn anything: 
\eq{H_0= \{\mathbb{E}[Bac] \le 0.5 \} \, .}
and the alternative hypothesis -- which corresponds to ``discovery'' -- is that the expected balanced accuracy is above $0.5$
\eq{H_1= \{\mathbb{E}[Bac] > 0.5 \} \, . }
Sample averages, such as the empirical balanced accuracy, show a random deviation around the true expectation. Concentration inequalities, such as the \emph{Hoeffding's inequality}, can be used to give upper bounds on probabilities of deviations between the empirical balanced accuracies $Bac$ and its expected value $\mathbb{E}[Bac]$\cite{hoeffdings}.\\

\emph{Hoeffding's inequality} states the following:
Let $\overline{X} = \frac{1}{n}\sum_{i=1}^n X_i$ be an empirical sample average. Furthermore the $X_i$ are independent (but not identically distributed) random variables with values $a_i \le X_i \le b_i$. Then a bound for the probability of excess deviation of the empirical average $\overline{X}$ over its expectation $\mathbb{E}[X]$ is given as:
\eq
{
P(\overline{X}- \mathbb{E}[X] \ge t) \le \exp\left( -\frac{2n^2 t^2}{\sum_{i=1}^n (b_i-a_i)^2}\right) \label{eq:hoeff}
}
The requirements for this tests are satisfied as scores of patients can be considered independent and we can find a lower and upper bound for $Bac$: $a_i=0$, and $b_i= \frac{n}{2n_+}$ for the $n_+$ many positive samples, and $b_i= \frac{n}{2n_-}$ for the $n_-$ many negative samples. By inserting the bounds into~\ref{eq:hoeff} we obtain:
\eq
{
P(\overline{X}- \mathbb{E}[X] \ge t) \le \exp\left( -2 t^2\frac{4 n_+ n_-}{n_+ + n_-}\right) \label{eq:hoeff2}
}
For a more detailed and step-by-step derivation, please refer to derivation at the end of this chapter. Now, if $\overline{X}$ is fixed, and if we consider different values for $\mathbb{E}[X]$, then $t= \overline{X}- \mathbb{E}[X]  $ increases as $\mathbb{E}[X]$ decreases. As a consequence the probability $\exp\left( -2 t^2\frac{4 n_+ n_-}{n_+ + n_-}\right)$ decreases. Therefore the probability bound from eq.~\ref{eq:hoeff2} is an increasing function of $t= \overline{X}- \mathbb{E}[X]$, and a decreasing function in $\mathbb{E}[X]$. This means that when plugging in $\mathbb{E}[X]=0.5$ in this inequality, we also cover the excess probabilities for the case when $\mathbb{E}[X] < 0.5$. This yields a statistical test for a single gene where the $p$-value for the test is given as
\eq{
p&=\exp\left( -2 t^2\frac{4 n_+ n_-}{n_+ + n_-}\right) = \exp\left( -2  \mathrm{max}(\overline{Bac}-0.5,0)^2 \frac{4 n_+ n_-}{n_+ + n_-}\right)\\
 &= \exp\left( - \mathrm{max}(\overline{Bac}-0.5,0)^2 \cdot 8q(1-q)n\right) \, , \label{eq:pvalue}
}
where $n$ is the number of samples used to compute the balanced accuracy, and $q \in \{0.1,\ldots,0.9\}$ is the quantile used for thresholding the true labels.

These $p$-values can now be applied in the Benjamini-Hochberg FDR test framework \newline\cite{bhprocedure}. This implies ordering the $p$-values $p_k$ across multiple experiments $k=1,\ldots,K$ and counting those until the condition
\eq{
p_k \le \alpha\frac{k}{K}, k=1,\ldots, K
}
is violated the first time.

Since training is based on correlation between the fixed kernel matrix and different sets of labels, which are partially overlapping, one can expect a positive correlation between balanced accuracies for most of the genes. Thus we used the variant of Benjamini-Hochberg for positive correlations. It should be noted, however, that assuming negative correlations for $K \approx 20000$ implies dividing the $p$-value threshold by $10.5$, and this would amount to rerun the same test with an FDR of $\alpha=\frac{0.05}{10.5} \approx 0.005$. In a preliminary test using $\alpha=0.005$ that there are still a large number of significant findings (e.g.~$2144$ for METH and $4936$ for CNV), if one would apply an assumption of negative correlation, which however is not justifiable to be used.

It should be noted, that the used bound, Hoeffdings inequality is a very \emph{conservative} estimate, as it does not use any information of the variances of single predictions. Thus the true discovery rate is expected to be much higher. A second test methodology described in the next Section confirms that.

\subparagraph{Derivation}
This derives the effective sample size for the Hoeffding bound.
\footnote{One can compare to the case of accuracy with $a_i=0,b_i=1$, where one would obtain a bound of $ \exp\left( -2 t^2n\right)$. The bound here for balanced accuracy $Bac$ yields an effective sample size $n_{eff}= \frac{4 n_+ n_-}{n_+ + n_-} = 4q(1-q)n$. This is sane, as for the median where balanced accuracy is equal to accuracy, one would have $n_+ = n_-  =0.5n$ and $ n_+ + n_-=n $, and thus an effective sample size of $n_{eff}=n$, whereas for the 10\%- and 90\%-quantiles it yields a reduced effective sample size of $n_{eff}= 4q(1-q)n = 4 \cdot 0.1 \cdot 0.9 n= 0.36n$.}
\eq{
\sum_{i=1}^n (b_i-a_i)^2 &= \sum_{i=1}^n 1[y_i=+1](b_i-a_i)^2 + 1[y_i=-1](b_i-a_i)^2\\
&=  \sum_{i=1}^n 1[y_i=+1](\frac{n}{2n_+}-0)^2 + 1[y_i=-1](\frac{n}{2n_-}-0)^2\\
&= \sum_{i=1}^n 1[y_i=+1]\frac{n^2}{4n_+^2} + 1[y_i=-1]\frac{n^2}{4n_-^2}\\
&= \sum_{i=1}^n 1[y_i=+1]\frac{n^2}{4n_+^2} + \sum_{i=1}^n1[y_i=-1]\frac{n^2}{4n_-^2}\\
&= n_+\frac{n^2}{4n_+^2} + n_-\frac{n^2}{4n_-^2}\\
&= \frac{n^2}{4n_+} + \frac{n^2}{4n_-}
}
Plugging this in into above equation yields
\eqn{
\exp\left( -\frac{2n^2 t^2}{\sum_{i=1}^n (b_i-a_i)^2}\right)&=  \exp\left( -\frac{2n^2 t^2}{\frac{n^2}{4n_+} + \frac{n^2}{4n_-}}\right) =  \exp\left( -\frac{2n^2 t^2}{\frac{n^2n_-}{4n_+n_-} + \frac{n^2n_+}{4n_+n_-}}\right)  \\
= \exp\left( -\frac{2 t^2 4 n_+ n_-}{n_+ + n_-}\right) & \label{eq:b1}
}
For a quantile $q \in [0,1]$ we have $n_+=nq$ and $n_-=n(1-q)$, $n_+ + n_- =n$ and thus
\eq{
= \exp\left( -\frac{2 t^2 4 nq \cdot n(1-q)}{n}\right) =  \exp\left( -8 t^2 nq(1-q)\right)  \label{eq:b2}
}

\subsection{Monte-Carlo estimation of the false discovery rate}

This test relies on Monte-Carlo simulation based on randomized labels from random permutations. The idea behind this is to compare the distributions of balanced accuracies computed with true labels and with randomized labels, utilizing the same learning approach. If the approach would have learned random structures, due to an uninformative kernel matrix or overfitting, then the two distributions would be very similar. Thus the null hypothesis $H_0$ states that both distributions are the same.

In practical terms this means that we use the same kernels, but instead of partitioning the values of the true labels, we partition random permutations of our $n$ samples into quantiles of 10\% to 90\%. Then we run the same learning experiments with over $K>1000$ independent experiments, each with simulated random labels, and obtain the distribution of balanced accuracies for the $K$ simulated experiments. If we consider the statistics, these simulations provide an estimate for the case of false discovery
\eq{P(Bac \ge t | False \, Discovery)}
while our experiments with true labels are an estimate for
\eq{P(Bac \ge t) =  P(Bac \ge t , False \, Discovery)  + P(Bac \ge t , \neg \, False \, Discovery)}
thus the sum of unconditional distribution of balanced accuracies with False and True Discoveries. From this we can obtain an estimate for the false discovery rate conditioned on balanced accuracies over a certain threshold $t$
\eq{
P(False \, Discovery |Bac \ge t ) = \frac{P(Bac \ge t | False \, Discovery) P(False \, Discovery)}{P(Bac \ge t)}  
}
As we have no estimate for $P(False \, Discovery)$, we bound it as: $$P(False \, Discovery) \le 1$$  
to obtain an upper bound on $P(False \, Discovery |Bac \ge t )$
\eq{
P(False \, Discovery |Bac \ge t )  &\le \min\left(1, \frac{P(Bac \ge t | False \, Discovery)}{P(Bac \ge t)}  \right) \\
&= \min\left(1, \frac{MC(Bac \ge t)}{True(Bac \ge t)}\right)
}
This is the quotient of the tail-cumulative distributions between the Monte-Carlo estimate and the distribution over the true labels.

The advantage of this approach is that it is highly similar to the test setup as we merely replace true labels by random permutations, and that it makes very little assumptions. However this approach does not provide decisions if single genes are significant. Furthermore this approach is computationally expensive, as it requires to run the same setup for many thousands of ``hallucinated by randomization'' genes.

Note that this test is very conservative: For many permutations, the labels obtained from random permutations for binary classification might be identical or highly overlapping to the labels obtained from true measurements. Therefore, when the random permutation is being close to true labels, one will obtain an empirical balanced accuracy greater than $0.5$ if the true labels are predictable. Further, our estimates based on the Hoeffdings inequality show that the estimation $P(False \, Discovery |Bac \ge t ) \leq 1$ is too conservative.

\end{document}